\documentclass{elsarticle}
\usepackage{framed,multirow}

\usepackage{amssymb}
\usepackage{latexsym}
\usepackage{subcaption}

\usepackage{bm}
\usepackage{amsmath}
\usepackage{graphicx}

\usepackage{adjustbox}
\usepackage{booktabs}
\usepackage{multirow}
\usepackage[table]{xcolor}% http://ctan.org/pkg/xcolor
\usepackage{lscape}
\usepackage{array}
\usepackage{graphicx}
\usepackage{geometry}
\usepackage{url}
\usepackage{bm}
\usepackage{xcolor}
\usepackage{float}
\usepackage{amsmath}
\usepackage{array}
\usepackage{pgfplots}

\usepackage{hyperref}

\usepackage[normalem]{ulem} %sout

\definecolor{bblue}{HTML}{4F81BD}
\definecolor{rred}{HTML}{C0504D}
\definecolor{ggreen}{HTML}{9BBB59}
\definecolor{ppurple}{HTML}{9F4C7C}
\definecolor{oorange}{HTML}{BD814F}
\definecolor{rrare}{HTML}{81BD4F}
\definecolor{ggray}{HTML}{BD4F81}

\usepackage{pgfplotstable} %transpose table
\usepackage{booktabs}
\usepackage{adjustbox}
\usepackage{multirow}
\usepackage{tikz}

\pgfplotsset{
/pgfplots/my legend/.style={
legend image code/.code={
\draw[thick,black](-0.05cm,0cm) -- (0.3cm,0cm);%
   }
  }
}
\usepackage{pifont}

\definecolor{newcolor}{rgb}{.8,.349,.1}

\usepackage{amsmath,amssymb,amsfonts}
\usepackage{algorithmic}
\usepackage{graphicx}
\usepackage{algorithm,algorithmic}
\usepackage{hyperref}
\hypersetup{hidelinks=true}
\usepackage{textcomp}
\usepackage{mathtools}
\usepackage{comment}

\usepackage{adjustbox}
\usepackage{multirow}
\usepackage{tikz}

\usepackage{bm}
\usepackage{lscape}
\usepackage{array}
\usepackage{geometry}

\usepackage{latexsym}
\usepackage{subcaption}
\usepackage{bbm}  % Include this to use \mathbbm

\newcommand{\R}{\mathbb{R}} %Real numbers
\newcommand{\E}{\mathbb{E}} %Expectation
\newcommand{\N}{\mathcal{N}} %gaussian normal
\newcommand{\M}{\mathcal{M}} %Conjunto
\newcommand{\const}{\text{const}} %Expectation
\DeclarePairedDelimiter\p{(}{)}

\DeclarePairedDelimiter\llav{\{}{\}}
\DeclarePairedDelimiter\ang{\langle}{\rangle}

\DeclareMathOperator{\yn}{y_{n,:}}

\DeclareMathOperator{\acerotau}{\alpha_{0}^{\tau}}
\DeclareMathOperator{\bcerotau}{\beta_{0}^{\tau}}
\DeclareMathOperator{\taus}{\tau}
\DeclareMathOperator{\tauE}{\langle\tau\rangle}

\DeclareMathOperator{\unmed}{\frac{1}{2}}
\DeclareMathOperator{\lnpi}{\ln(2\pi)}
\DeclareMathOperator{\lntau}{\ln(\tau)}

\DeclareMathOperator{\Am}{\mathbf{A}^{(m)}} %\mathbf{w}_:k^(m)
\DeclareMathOperator{\AmT}{\mathbf{A}^{(m)^\top}} %\mathbf{w}_:k^(m)

\DeclareMathOperator{\sumn}{\sum\limits_{n=1}^{N}} %sum_{n=1}^{N}
\DeclareMathOperator{\summ}{\sum\limits_{m=1}^{M}} %sum_{m=1}^{2}
\DeclareMathOperator{\sumk}{\sum\limits_{k=1}^{K}} %sum_{k=1}^{K_c}
\DeclareMathOperator{\sumd}{\sum\limits_{d=1}^{D}} %sum_{d=1}^{D_m}
\DeclareMathOperator{\sumc}{\sum\limits_{c=1}^{C}} %sum_{n=1}^{N}
\DeclareMathOperator{\Ik}{\mathbf{I}_{K}} %W^\top
\DeclareMathOperator{\Ic}{\mathbf{I}_{C}} %W^\top

\newcommand{\eqline}{\enskip&\enskip}
\newcommand{\eqeq}{\enskip=&\enskip}

\newcommand{\eqsimil}{\enskip\sim &\enskip}

\DeclareMathOperator{\y}{\mathbf{y}} %X

\DeclareMathOperator{\Z}{\mathbf{Z}} %Y
\DeclareMathOperator{\ZT}{\mathbf{Z}^\top} %Y^\top
\DeclareMathOperator{\Zn}{\mathbf{z}_{n,:}} %Y
\DeclareMathOperator{\ZnT}{\mathbf{z}_{n,:}^\top} %Y^\top
\DeclareMathOperator{\Wk}{\mathbf{w}_{k,:}} %W^(m)
\DeclareMathOperator{\Wkd}{w_{k,d}^{(m)}} %W^(m)
\DeclareMathOperator{\X}{\mathbf{X}} %W^(m)
\DeclareMathOperator{\Xn}{\mathbf{x}_{n,:}} %W^(m)
\DeclareMathOperator{\Y}{\mathbf{Y}} %W^(m)
\DeclareMathOperator{\Yn}{\mathbf{y}_{n,:}} %W^(m)
\DeclareMathOperator{\YT}{\mathbf{Y}^\top} %W^(m)
\DeclareMathOperator{\YnT}{\mathbf{y}_{n,:}^\top} %W^(m)
\DeclareMathOperator{\V}{\mathbf{V}} %W^(m)
\DeclareMathOperator{\Vc}{\mathbf{v}_{c,:}} %W^(m)
\DeclareMathOperator{\Vk}{\mathbf{v}_{:,k}} %W^(m)
\DeclareMathOperator{\VT}{\mathbf{V}^\top} %W^(m)
\DeclareMathOperator{\VcT}{\mathbf{v}^\top_{c,:}} %W^(m)
\DeclareMathOperator{\VkT}{\mathbf{v}^\top_{:,k}} %W^(m)

\DeclareMathOperator{\bet}{\text{\boldmath$\omega$}} %alpha^(m)
\DeclareMathOperator{\gamd}{\gamma_d^{(m)}} %alpha_k^(m)

\DeclareMathOperator{\Xm}{\mathbf{X}^{(m)}} %X^(m)
\DeclareMathOperator{\XmT}{\mathbf{X}^{(m)^\top}} %X^(m)^\top
\DeclareMathOperator{\Xnm}{\mathbf{x}_{n,:}^{(m)}} %X_n^(m)
\DeclareMathOperator{\XnmT}{\mathbf{x}_{n,:}^{(m)^\top}} %X_n^(m)^\top
\DeclareMathOperator{\Wm}{\mathbf{W}^{(m)}} %W^(m)
\DeclareMathOperator{\WmT}{\mathbf{W}^{(m)^\top}} %W^(m)^\top
\DeclareMathOperator{\Wkm}{\mathbf{w}_{k,:}^{(m)}} %\mathbf{w}_:k^(m)
\DeclareMathOperator{\akm}{\alpha_k^{(m)}} %alpha_k^(m)
\DeclareMathOperator{\gamm}{\text{\boldmath$\gamma$}^{(m)}} %gamma^(m)
\DeclareMathOperator{\Xsm}{\mathbf{\tilde{X}}^{(m)}} 
\DeclareMathOperator{\Xsmt}{\mathbf{\tilde{X}}^{(m)\top}}

\DeclareMathOperator{\tpredc}{t_{\ast,c}}

\DeclareMathOperator{\ypredc}{y_{\ast,c}}
\DeclareMathOperator{\zpred}{\mathbf{z}_{\ast,:}}
\DeclareMathOperator{\zpredT}{\mathbf{z}_{\ast,:}^\top}
\DeclareMathOperator{\xpredin}{\mathbf{x}_{\ast,:}^{\M}}

\journal{arXiv}

\begin{document}

% \verso{Albert Belenguer-Llorens \textit{et~al.}}

\begin{frontmatter}

\title{Unified Bayesian representation for high-dimensional multi-modal biomedical data for small-sample classification}
% The Relevance Feature and Vector Machine for Health Applications
%

\author[1]{Albert Belenguer-Llorens\corref{cor1}}
\cortext[cor1]{Corresponding author: 
  Tel.: +0-000-000-0000;}
\ead{abelengu@pa.uc3m.es}

\author[2]{Carlos Sevilla-Salcedo}
\author[1]{Jussi Tohka}
\author[1,3]{Vanessa G\'{o}mez-Verdejo}
\author{for the Alzheimer's Disease Neuroimaging Initiative}

\address[1]{Department of Signal Processing and Communications, Universidad Carlos III de Madrid, Leganés, 28911, Spain}
\address[2]{Department of Computer Science, Aalto University, Espoo, 02150, Finland}
\address[3]{Instituto de Investigación Sanitaria Gregorio Marañón (IiSGM), Madrid, 28009 Spain}

% \received{1 May 2013}
% \finalform{10 May 2013}
% \accepted{13 May 2013}
% \availableonline{15 May 2013}
% \communicated{S. Sarkar}

\begin{abstract}
%%%

We present BALDUR, a novel Bayesian algorithm designed to deal with multi-modal datasets and small sample sizes in high-dimensional settings while providing explainable solutions. To do so, the proposed model combines within a common latent space the different data views to extract the relevant information to solve the classification task and prune out the irrelevant/redundant features/data views. Furthermore, to provide generalizable solutions in small sample size scenarios, BALDUR efficiently integrates dual kernels over the views with a small sample-to-feature ratio. Finally, its linear nature ensures the explainability of the model outcomes, allowing its use for biomarker identification. This model was tested over two different neurodegeneration datasets, outperforming the state-of-the-art models and detecting features aligned with markers already described in the scientific literature.

\end{abstract}


\end{frontmatter}

%\linenumbers

%% main text

\section{Introduction}
\label{sec:Intro}

The rapid digitization and advances in healthcare technologies have led to the generation of large, heterogeneous datasets, encompassing various modalities such as medical imaging, genetic information, and blood measurements. Effectively combining these diverse, complex, and heterogeneous data sources to extract meaningful insights remains a major challenge for Machine Learning (ML) algorithms \cite{acosta2022multimodal}. Working with multi-modal data becomes even more challenging when dealing with neuroimaging \cite{zhang2020advances} mainly due to the redundancy between modalities. Also, the contextual dependency in neuroimaging, where each voxel's significance is derived from its relationship to others, adds a new layer of complexity when defining ML algorithms. This dependency complicates the immediate concatenation of diverse modalities. This issue is even more complex in scenarios with small sample sizes and large feature sets. This condition, often referred to as wide data, presents computational and learning challenges as models fail to identify meaningful patterns, leading to unreliable and non-generalizable solutions \cite{konietschke2021small,bacchetti2013small}.

Given these challenges, it is essential to develop tailored ML algorithms that operate efficiently in multi-modal environments with small sample sizes and large feature sets. Additionally, since these models are intended to work in medical settings, explainability is crucial \cite{ratti2022explainable}. Identifying which medical tests or variables the model relies on for diagnostic decisions plays a key role in building trust with clinicians and uncovering potential biomarkers \cite{rasheed2022explainable}.

To address these limitations, we propose a novel algorithm, the \textbf{BA}yesian \textbf{L}atent \textbf{D}ata \textbf{U}nified \textbf{R}epresentation (BALDUR). BALDUR efficiently combines different data sources, both wide and non-wide, by projecting all data views into a common latent space by using a Bayesian formulation. Furthermore, the model performs Feature Selection (FS) by imposing sparsity over the feature space while using kernelized representations over the wide views to enhance the model generalization and avoid the overfitting often derived from working with small sample sizes. Additionally, its linear formulation provides the basis for an explainable model that can identify and justify its decision-making for clinicians.

\section{Related work}

Related work in ML algorithms capable to integrate multi-modal data can be divided into two types of methods: classical ML and Deep Learning (DL) methods. Regarding classical ML methods, the state-of-the-art emphasizes Multiple Kernel Learning Support Vector Machines (MKL-SVM) \cite{gonen2011multiple}, which are commonly applied to overcome challenges in high-dimensional and multi-modal biomedical data \cite{zhang2024explaining} often citing their simplicity and explainability as the main advantages. However, MKL-SVM lacks joint FS across views and generates the latent representation in an unsupervised manner, meaning it does not tailor the latent space to the specific task. Beyond MKL-SVM, various models aim to combine data from multiple sources into a latent representation. Recent approaches based on weighted linear combinations \cite{yang2019adaptive, tao2018multiview} often struggle with high-dimensional data (especially with small sample sizes), due to the large number of model parameters to adjust. Self-representation-based algorithms are also becoming popular in biomedicine due to their ability to generate complex data combinations, surpassing traditional linear models \cite{luo2018consistent, lan2021generalized, li2019reciprocal}; however, they still struggling in wide data due to the huge complexity imposed by the large number of parameters to infer. Spectral clustering approaches \cite{liang2019consistency, kang2020partition} highlight as potential methods to carry out these constraints. These models integrate graph learning, spectral clustering, and partition fusion to iteratively optimize a final clustering process allowing them to handle high-dimensional data without computational limitations \cite{singh2020study,zheng2022multi}. Nonetheless, this hierarchical approach adds noise and redundancy over the generated spectral space, as it does not have prior information of the following task.

On the other hand, DL methods have also been applied to combine data from different sources, typically by processing each view through an embedding network and combining them into a common latent space. Many medical applications of DL-based models can be found across the literature, with different architectures and strategies \cite{pei2023review,behrad2022overview,ravi2016deep}. However, DL-based models face two main limitations: (1) Due to their complexity, these models tend to overfit when working with small sample cohorts (especially when handling high-dimensional data), and (2), due to their non-linear nature, they are unable to provide explainable solutions. Thus, these constraints make them unsuitable for defining reliable clinical tools for high-dimensional contexts.

\section{The proposed model}

\subsection{Generative Model}
\label{sec:generative}

To overcome the above limitations previously discussed in Section \ref{sec:Intro}, we here introduce \textbf{BA}yesian \textbf{L}atent \textbf{D}ata \textbf{U}nified \textbf{R}epresentation (BALDUR). This model builds on the foundations of the Sparse Bayesian Partial Least Squares  \cite{vidaurre2013bayesian}, but includes capabilities to handle multi-modal data while efficiently working in wide data environments.

Let us define a regression dataset composed of $N$ observations and their corresponding targets, denoted as $\{(\Xn^{\M}, \Yn)\}_{\rm n=1}^N$, were we define a set of $ \M=\{1,2,\dots,M\}$ views and each $\Xn^{\rm (m)} \in \R^{D^{\rm (m)}}$ represents the $n$-th observation of the $m$-th view, composed of $D^{\rm (m)}$ elements. Also, $\Yn \in \R^{C}$ denotes the $n$-th observation`s $C$ targets.

Our model, establishes a relationship between the input sample $\Xn^{\M}$ and its corresponding regression target $\Yn$ through a latent variable $\Zn$ with a two-step hierarchical approach as
\begin{align}
    \Zn &= \sum_m^M\left(\Xnm \mathbf{W}^{\rm (m)\top}\right) + {\boldsymbol \epsilon}_{\rm \mathbf Z} \label{eq:zz}\\
    \Yn &= \Zn \VT + \; {\boldsymbol \epsilon}_{\rm \mathbf Y} \label{eq:y_SBPLS},
\end{align}
where ${\boldsymbol \epsilon}_{\rm \mathbf Z}$ and ${\boldsymbol \epsilon}_{\rm \mathbf Y}$ are Gaussian noises with zero mean and precision following a Gamma distribution as $\tau \sim \Gamma\p*{\alpha^{\tau}, \beta^{\tau}}$ and $\psi \sim \Gamma\p*{\alpha^{\psi}, \beta^{\psi}}$, i.e., ${\boldsymbol \epsilon}_{\rm \mathbf Z} \sim \N(0,\tau^{-1}\Ik)$ and ${\boldsymbol \epsilon}_{\rm \mathbf Y} \sim \N(0,\psi^{-1}\Ic)$. %Moreover, $\mathbbm{1}\left(s^{\rm (m)} = 0\right)$ is an indicator function that equals 1 if $s^{\rm (m)} = 0$. 

Furthermore, to impose sparsity on the specific contribution of each data view over the latent factors, we define the weight vectors as
\begin{equation} 
    \Wkd  \sim \N\p*{0,\p*{\delta_{\rm k}^{\rm (m)} \gamd}^{-1}}, \label{eq:Wkd}
\end{equation}
where $\delta_{\rm k}^{\rm (m)} \sim \Gamma\p*{\alpha^{\delta^{\rm (m)}}_{\rm k}, \beta^{\delta^{\rm (m)}}_{\rm k}}$ and $\gamd \sim \Gamma\p*{\alpha^{\gamma^{\rm (m)}}_{\rm d}, \beta^{\gamma^{\rm (m)}}_{\rm d}}$ which defines a double Automatic Relevance Determination (ARD) prior over features and latent factors. This way, the model sets sparsity over the feature weights while defining common and specific latent factors per view. Moreover, to impose sparsity in multi-output settings, we define $\Vk  \sim \N\p*{0,\omega_{\rm k}^{-1}\Ic}$ where $\omega_{\rm k} \sim \Gamma\p*{\alpha^{\omega}_{\rm k}, \beta^{\omega}_{\rm k}}$.

\begin{figure*}[ht!]
    \centering
    % First row
    \begin{subfigure}[b]{0.32\textwidth}
        \centering
        \includegraphics[width=\textwidth]{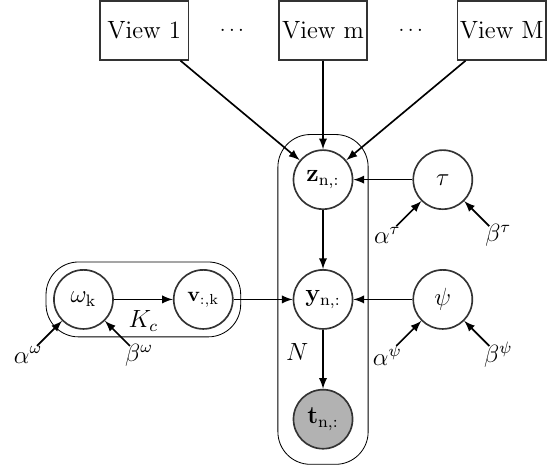} 
        \caption{BALDUR model}
    \end{subfigure}
    % Second row
    % \hskip\baselineskip
    \begin{subfigure}[b]{0.32\textwidth}
        \centering
        \includegraphics[width=\linewidth]{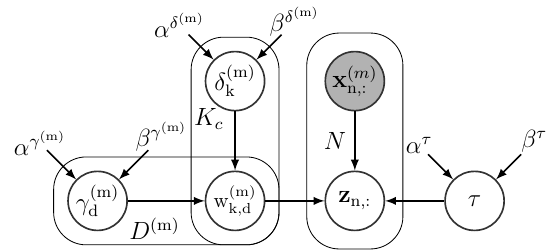}
        \caption{Primal view}
    \end{subfigure}
    % \hskip\baselineskip
    \begin{subfigure}[b]{0.32\textwidth}
        \centering
        \includegraphics[width=\linewidth]{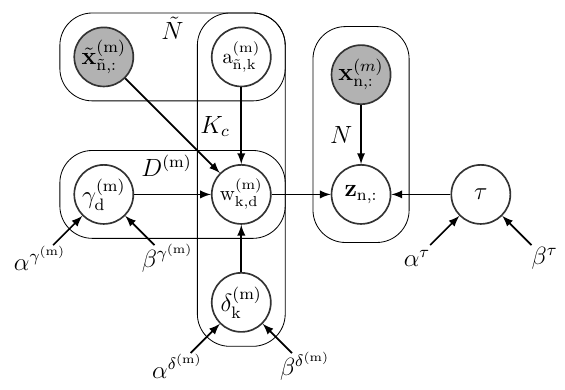}
        \caption{Dual view}
    \end{subfigure}

    \caption{Diagram of the graphical model of BALDUR for classification tasks (top) and the two possible view settings: primal (middle) and dual (bottom). Grey circles denote observed variables, white circles unobserved random variables and rectangles represent node groups dependent on the view's feature space. The nodes without a circle correspond to the hyperparameters.}
    \label{fig:GM_BALDUR}
\end{figure*}

Additionally, to allow BALDUR to efficiently work either with normal and wide data, we included dual variables $\mathbf{A}^{(m)}$ in Eq. (\ref{eq:zz}) following the idea presented in \cite{belenguer2022novel} as:
\begin{equation}
\begin{split}
    \Zn & = \sum_m^M\Bigg[\left(\Xnm \mathbf{W}^{\rm (m)\top}\right)\mathbbm{1}\left(s^{\rm (m)} = 0\right) \nonumber \\ &+ \left(\Xnm \mathbf{\tilde{X}}^{(m)\top}\mathbf{A}^{\rm (m)}\right)\mathbbm{1}\left(s^{\rm (m)} = 1\right)\Bigg] + {\boldsymbol \epsilon}_{\rm \mathbf Z},
\end{split}
\end{equation}
where $\mathbbm{1}\left(s^{\rm (m)} = 0\right)$ is an indicator function that equals 1 if $s^{\rm (m)} = 0$, indicating that the $m$-th view is not wide (see Figure \ref{fig:GM_BALDUR}(b)), and 0 otherwise (see Figure \ref{fig:GM_BALDUR}(c)). Besides, Figure \ref{fig:GM_BALDUR}(a) depicts the general model of BALDUR.

Note that this variable establishes a linear relation with $\mathbf{W}^{(m)}$ following $\mathbf{W}^{(m)\top} = \mathbf{\tilde{X}}^{(m)\top}\mathbf{A}^{(m)}$, where $ \mathbf{\tilde{X}}^{(m)} \in \R^{\tilde{N} \times D^{(m)}}$ is a subset of the training data points of the $m$-th view used to represent the solution, known as Relevance Vectors (RV), and $\mathbf{A}^{(m)} \in \R^{\tilde{N} \times K}$ its associated dual variables. Also, unlike Relevance Vector Machines (RVM) \cite{tipping1999relevance}, our model infers the posterior of $\bf A$ while keeping the sparsity over the features using the ARD, i.e., we did not set a prior distribution over $\mathbf{A}^{(m)}$ but we parameterize its posterior during the further inference. Hence, the model will be able to overcome the wide data limitations as it will recursively calculate a ($\tilde{N},\tilde{N}$)-dimensional covariance matrix instead of ($D^{(m)},D^{(m)}$)-dimensional.

%%%%%%%%%%%%%%%
%%%%%%%%%%%%%%%

Finally, as the model is intended to solve classification problems, we will draw on the Bayesian logistic regression formulation presented in \cite{jaakkola1997variational}. We now consider the output view to be multi-labeled, so that the binary data $\mathbf{t}_{\rm n,:} \in \R^{C}$ is observed, and $\mathbf{y}_{\rm n,:}$ is now an unobserved random variable. Thus, we will define the relation between the regression output $\mathbf{y}_{\rm n,:}$ and the output label $\mathbf{t}_{\rm n,:}$ through a Bayesian logistic regression as
\begin{equation}
\label{eq:reg_mod}
    p(\mathbf{t}_{\rm n,:}|\mathbf{y}_{\rm n,:}) = e^{\mathbf{y}_{\rm n,:} \mathbf{t}_{\rm n,:}}\sigma(-\mathbf{y}_{\rm n,:}),
\end{equation}
where $\sigma(x) = (1+e^{-x})^{-1}$ is the sigmoid function.

\subsection{Model Inference}
\label{sec:Inference_BALDUR}

Once the generative model and its prior distributions are defined, we need to parametrize the posterior distribution of the model variables using the observed data, $p(\bf \boldsymbol{\Theta}| \mathbf{X}^{\M}, \mathbf{t})$ where $\boldsymbol{\Theta}$ contains the model parameters. However, defining an analytical expression for the posterior distribution parameters is infeasible. To overcome this challenge, we approximate the posterior distribution $p(\boldsymbol{\Theta}| \mathbf{X}^{\M}, \mathbf{t})$ using the mean-field variational inference procedure \cite{blei2017variational} to a new factorized posterior $q(\boldsymbol{\Theta})$ that assumes independence between the model variables as
\begin{equation}
\begin{split}
    &p(\boldsymbol{\Theta}| \mathbf{X}^{\M}, \mathbf{t})  \backsimeq   q(\boldsymbol{\Theta}) =  \prod_{m}^{M}\Bigg(q\left(\Wm\right)^{\mathbbm{1}(s^{\rm (m)} = 0)} \\ &q\left(\mathbf{A}^{\rm (m)}\right)^{\mathbbm{1}(s^{\rm (m)} = 1)}\prod_{\rm k}^{K}q\left(\delta_{\rm k}^{(m)}\right)\prod_{\rm d}^{D}q\left(\gamma_{\rm d}^{(m)}\right)\Bigg) \\ &\prod_n^{N}\left(q\left(\mathbf{z}_{\rm n,:}\right)q\left(\mathbf{y}_{\rm n,:}\right)\right)q(\mathbf{V})\prod_k^K\left(q(\omega_{\rm k})\right)q(\tau)q(\psi),
\end{split}
\end{equation}
where $q(\cdot)$ represents the approximated posterior distribution of a variable. Also, to measure the quality of the approximation $q(\bf \Theta)$, the mean-field approximation maximizes a lower bound $L(q)$, proportional to the Kullback-Leibler (KL) divergence between $q(\bf \Theta)$ and $p(\bf \Theta| \mathbf{X}^{\M}, \mathbf{t})$, defined as
\begin{equation}
    L(q) = \E_{q}\left[\ln (q(\bm{\Theta}))\right] - \E_{q}\left[\ln (p(\bm{\Theta}, \mathbf{t}, \mathbf{X}^{\M}))\right],
    \label{eq:elboo}
\end{equation}
where $\E_{q}\left[\ln (p(\bm{\Theta}, \mathbf{t}, \mathbf{X}^{\M}))\right]$ is the expectation of the joint posterior distribution w.r.t. $q(\mathbf{\Theta})$ and $\E_{q}[\ln 
 q(\bm{\Theta})]$ the entropy of $q(\mathbf{\Theta})$.

Besides, as presented in \cite{bishop2006pattern}, by analysing the expression $L(q)$ we can compute the approximated posterior of any $\theta_j$ of the model following:
\begin{equation} 
    \ln q_j^* = \E_{-q_j}\left[\ln (p(\bm{\Theta}, \mathbf{t}, \mathbf{X}^{\M}))\right] + const, 
\label{eq:lnqopt}
\end{equation} 
where $\E_{-q_j}\left[\cdot\right]$ refers to the expectation over all the r.v. expect the $j$-th. Also, as Eq. (\ref{eq:lnqopt}) requires the distributions to be conjugated, we have to slightly modify Eq. \eqref{eq:reg_mod} following the idea presented in \cite{jaakkola2000bayesian}. That is, we lower-bound $p(t_{\rm n,c}|y_{\rm n,c})$ based on first-order Taylor series expansion as
\begin{equation}
\label{eq:lbb}
\begin{split}
    & p(t_{\rm n,c} = 1|y_{\rm n,c}) = 
    e^{y_{\rm n,c}}\sigma(-y_{\rm n,c}) \geqslant h(y_{\rm n,c}, \xi_{\rm n,c}) \\ &= 
    e^{y_{\rm n,c} t_{\rm n,c}}\sigma(\xi_{\rm n,c})e^{-\frac{y_{\rm n,c} + \xi_{\rm n,c}}{2} - \lambda(\xi_{\rm n,c})(y_{\rm n,c}^2 - \xi_{\rm n,c}^2)},
\end{split}
\end{equation}
being $\lambda(a) = \frac{1}{2a}(\sigma(a) - \unmed)$ and $\xi_{\rm n,c}$ the center of the Taylor series around $y_{\rm n,c}$ (see \ref{sec:Ap_logbound} for futher mathematical developments). This trick allows us to treat $p(t_{\rm n,c}|y_{\rm n,c})$ as a Gaussian distribution and subsequently being introduced in Eq. \eqref{eq:lnqopt}. 

Table \ref{tab:my_label} shows the model mean-field expression update rules obtained following Eq. \eqref{eq:lnqopt} (the detailed mathematical developments of these expressions are depicted in \ref{sec:Ap_inf_BALDUR}). Also, if we apply Eq. (\ref{eq:elboo}), we can depict the final $L(q)$ expression as:
\begin{equation}
\label{eq:Lq}
    \begin{split}
        L(q) = &\frac{\N}{2}\ln |\Sigma_{\mathbf{Z}}|- (2 + \frac{\N}{2} - \alpha_0^\tau)\ln(\beta^\tau) + \frac{\N}{2}\ln|\Sigma_{\mathbf{y}}| + \frac{C}{2}\ln|\Sigma_{\mathbf{V}}| + (\alpha_0^\omega -2)\sum_{\rm k}^K\ln(\beta_{\rm k}^\omega) \\ & - (2 + \frac{\N}{2} - \alpha_0^\psi)\ln(\beta^{\psi}) + \sum_m^M\bigg[\left(\frac{K}{2} + \alpha_0^{\gamma^{\rm (m)}} -2\right)\sum_{\rm d}^{D^{\rm (m)}}\ln(\beta_{\rm d}^{\gamma^{\rm (m)}}) \\ & + \left(\frac{K}{2} + \alpha_0^{\delta^{\rm (m)}} -2\right)\sum_{\rm k}^K\ln(\beta_{\rm k}^{\delta^{\rm (m)}}) + \sum_{\rm d}^{D^{\rm (m)}}\left(\beta_0^{\gamma^{\rm (m)}}\frac{\alpha_{\rm d}^{\gamma^{\rm (m)}}}{\beta_{\rm d}^{\gamma^{\rm (m)}}}\right) + \sum_{\rm k}^K\left(\beta_0^{\delta^{\rm (m)}}\frac{\alpha_{\rm k}^{\delta^{\rm (m)}}}{\beta_{\rm k}^{\delta^{\rm (m)}}}\right) \\ & + \unmed\sum_{\rm k}^K\sum_{\rm d}^{D^{\rm (m)}}\left(\frac{\alpha_{\rm k}^{\delta^{\rm (m)}} \alpha_{\rm d}^{\gamma^{\rm (m)}}}{\beta_{\rm k}^{\delta^{\rm (m)}} \beta_{\rm d}^{\gamma^{\rm (m)}}}\langle\Wkd\rangle\right)\mathbbm{1}(s^{\rm (m)} = 0) \\ & + \unmed\sum_{\rm k}^K\sum_{\rm d}^{D^{\rm (m)}}\left(\frac{\alpha_{\rm k}^{\delta^{\rm (m)}} \alpha_{\rm d}^{\gamma^{\rm (m)}}}{\beta_{\rm k}^{\delta^{\rm (m)}} \beta_{\rm d}^{\gamma^{\rm (m)}}}Tr\left(\Xsm\Xsmt\langle\mathbf{a}_{:,\rm k}\mathbf{a}_{:,\rm k}^T\rangle\right)\mathbbm{1}(s^{\rm (m)} = 1)\right) \bigg] \\ & - \sum_n^N\left(\ln(\sigma(\xi_{\rm n,:})) + \langle\mathbf{y}_{\rm n,:}\rangle\mathbf{t}_{\rm n,:}^\top - \unmed(\langle\mathbf{y}_{\rm n,:}\rangle + \xi_{\rm n,:}) - \lambda(\xi_{\rm n,:})(\langle\mathbf{y}_{\rm n,:}^2\rangle - \xi_{\rm n,:}^2)\right).
    \end{split}
\end{equation}

For further mathematical developments of this expression see \ref{sec:LB_BALDUR}.

\begin{table*}[ht!]
    \begin{adjustbox}{max width=\textwidth}
    \renewcommand{\arraystretch}{2}
        \centering
        \setlength{\tabcolsep}{1pt}
        \begin{tabular}{c c c}
        \hline
        {\bf Variable} & $\mathbf{q^{*}}$ {\bf distribution} & {\bf Parameters}\\
        \hline
        $a_{:,\rm k}^{\rm (m)}$ & $\N(a_{:,\rm k}^{\rm (m)}|\langle a_{:,\rm k}^{\rm (m)}\rangle,\Sigma_{a_{:,\rm k}}^{\rm (m)})$ & \begin{tabular}{@{}c@{}}$\Sigma_{a_{:,\rm k}}^{(m)-1} = \tauE \Xsm\XmT\Xm\Xsmt + \langle\akm \rangle\Xsm \Lambda_{\langle\gamm\rangle}\Xsmt$
        \\ $\langle\mathbf{a}_{:,\rm k}^{\rm (m)}\rangle = \tauE\Sigma_{a_{:,\rm k}}^{\rm (m)}\Xsm\XmT\left[\langle\mathbf{z}_{:,\rm k}\rangle - \sum_{m' \neq m}^{M}\mathbf{H}(m',s^{\rm (m')})_{:,k}\right]
        
        $\end{tabular}\\
        \hline
        $\Wkm$ & $\N(\Wkm|\langle \Wkm\rangle,\Sigma_{\Wk}^{\rm (m)})$ & \begin{tabular}{@{}c@{}}$\Sigma_{\Wk}^{(m)-1} = \tauE \XmT\Xm + \langle\akm \rangle \Lambda_{\langle\gamm\rangle}$
        \\ $\langle \Wkm\rangle = \tauE \left[\langle\mathbf{z}_{:,\rm k}\rangle - \sum_{m' \neq m}^{M}\mathbf{H}(m',s^{\rm (m')})^\top\right]\Xm\Sigma_{\Wk}^{\rm (m)}$\end{tabular}\\
        \hline
        $\delta_{\rm k}^{\rm (m)}$ & $\Gamma(\delta_{\rm k}^{\rm (m)}|\alpha_{\rm k}^{\delta(m)},\beta_{\rm k}^{\delta(m)})$ & \begin{tabular}{@{}c@{}}$\boldsymbol{\alpha}^{\delta(m)} = \frac{D^{\rm (m)}}{2} + \alpha_0^{\delta(m)}$ \\ $\beta_{\rm k}^{\delta(m)} = \beta_0^{\delta(m)} + \unmed\sum_{\rm d}^{D^{\rm (m)}}\left[\langle\gamma_d^{\rm (m)}\rangle\left(\langle w_{\rm k,d}^{(m)^{2}}\rangle\mathbbm{1}(s^{\rm (m)} = 0) + \left(\mathbf{\tilde{X}}_{:,\rm d}^{(m)\top}\langle\mathbf{a}_{:,\rm k}^{\rm (m)}\mathbf{a}_{:,\rm k}^{(m)\top}\rangle\tilde{X}_{:,\rm d}^{\rm (m)}\right)\mathbbm{1}(s^{\rm (m)} = 1)\right)\right]$\end{tabular}\\
        \hline
        $\gamma_{\rm d}^{\rm (m)}$ & $\Gamma(\gamma_{\rm d}^{\rm (m)}|\alpha_{\rm d}^{\gamma(m)},\beta_{\rm d}^{\gamma(m)})$ & \begin{tabular}{@{}c@{}}$\boldsymbol{\alpha}^{\gamma(m)} = \frac{K}{2} + \alpha_0^{\gamma(m)}$ \\ $\beta_{d\rm }^{\gamma(m)} = \beta_0^{\gamma(m)} + \unmed\sum_{\rm k}^{K}\left[\langle\delta_{\rm k}^{\rm (m)}\rangle\left(\langle w_{\rm k,d}^{(m)^{2}}\rangle\mathbbm{1}(s^{\rm (m)} = 0) + \left(\mathbf{\tilde{X}}_{:,\rm d}^{(m)\top}\langle\mathbf{a}_{:,\rm k}^{\rm (m)}\mathbf{a}_{:,\rm k}^{(m)\top}\rangle\tilde{X}_{:,\rm d}^{\rm (m)}\right)\mathbbm{1}(s^{\rm (m)} = 1)\right)\right]$\end{tabular}\\
        \hline
        $\mathbf{Z}$ & $\N(\mathbf{Z}|\langle \mathbf{Z}\rangle,\Sigma_{\mathbf{Z}})$ & \begin{tabular}{@{}c@{}}$\Sigma_{\mathbf{Z}}^{-1} = \tauE \Ik + \langle\psi\rangle\langle\mathbf{V}\mathbf{V}^\top\rangle$
        \\ $\langle\mathbf{Z}\rangle = \left(\langle\tau\rangle\sum_m^M \mathbf{H}(m,s^{\rm (m)}) + \langle\psi\rangle\mathbf{Y}\langle\mathbf{V}\rangle\right)\Sigma_{\mathbf{Z}}$\end{tabular}\\
        \hline
        $\tau$ & $\Gamma(\tau|\alpha^{\tau},\beta^{\tau})$ & \begin{tabular}{@{}c@{}} $\alpha^{\tau} = \frac{NK}{2} + \alpha_0^{\tau}$\\ $\beta^{\tau} = \beta_0^{\tau} + \unmed\sum_n^N\langle\mathbf{Z}\mathbf{Z}^\top\rangle_{n,n} - Tr\left(\langle\mathbf{Z}\rangle\sum_m^{M}\mathbf{H}(m,s^{\rm (m)})^\top\right) + \unmed Tr\left(\sum_m^M\sum_{m'}^M \mathbf{H}(m,s^{\rm (m)})\mathbf{H}(m',s^{(m')})^\top\right)$ \end{tabular}\\ 
        \hline
        $\mathbf{V}$ & $\N \p*{\V | \mu_{\V}, \Sigma_{\V}}$ & \begin{tabular}{@{}c@{}}$\Sigma_{\V}^{-1} = \Lambda_{\ang{\bet}} + \ang{\psi}\ang{\ZT\Z}$ \\ $\mu_{\V} = \ang{\psi} \YT \ang{\Z} \Sigma_{\V}$\end{tabular}\\
        \hline
        $\omega_{\rm k}$ & $\Gamma \p*{\omega_{\rm k} | \alpha_{\rm k}^{\omega}, \beta_{\rm k}^{\omega}}$ & \begin{tabular}{@{}c@{}}$\alpha_{\rm k}^{\omega} = \frac{C}{2} + \alpha^{\omega}_0$ \\ $\beta_{\rm k}^{\omega} = \beta^{\omega}_0 + \frac{1}{2}\ang{\VkT\Vk}$\end{tabular}\\
        \hline
        $\psi$ & $\Gamma \p*{\psi | \alpha^{\psi}, \beta^{\psi}}$ & \begin{tabular}{@{}c@{}}$\alpha^{\psi} = \frac{N C}{2} + \alpha^{\psi}_0$ \\ $\beta^{\psi} = \beta^{\psi}_0 + \frac{1}{2}\sum_n^N\mathbf{y}_{\rm n,:}\YnT - Tr\llav{\Y  \ang{\V} \ang{\ZT}} + \frac{1}{2} Tr\llav{\ang{\VT\V} \ang{\ZT\Z}}$\end{tabular}\\
        \hline
        $\mathbf{y}_{\rm n,:}$ & $\N(\mathbf{y}_{\rm n,:}|\langle \mathbf{y}_{\rm n,:}\rangle,\Sigma_{\mathbf{y}_{\rm n,:}})$ & \begin{tabular}{@{}c@{}c@{}c@{}}$\Sigma_{\mathbf{y}_{\rm n,:}}^{-1} = \langle\psi\rangle\Ic + 2\Lambda_{\xi_{\rm n,:}}$ \\ $\langle\mathbf{y}_{\rm n,:}\rangle = \left(\mathbf{t}_{\rm n,:} - \unmed + \langle\psi\rangle\langle\mathbf{z}_{\rm n,:}\rangle\langle\mathbf{V}^\top\rangle\right)\Sigma_{\mathbf{y}_{\rm n,:}}$ \\ where \\ $\xi_{\rm n,:} = \sqrt{\langle\mathbf{y}_{\rm n,:}\rangle^2 + \Lambda_{\Sigma_{\mathbf{y}_{\rm n,:}}}}$ \end{tabular}\\
        \hline
        \end{tabular}
    \end{adjustbox}
    \caption{$q^{*}$ update rule for all r.v.obtained using the mean-field approximation. Also, note that $\mathbf{H}(m,s^{\rm (m)}) =  \Xm\langle\Wm\rangle^\top$ if $s^{\rm (m)}=0$ or $\mathbf{H}(m,s^{\rm (m)}) =  \Xm\Xsmt\langle\mathbf{A}^{\rm (m)}\rangle$ if $s^{\rm (m)}=1$.}
    \label{tab:my_label}
\end{table*}

\subsection{Posterior predictive distribution}

To calculate the predictive distribution of the model we have to marginalize the posterior distribution of $\mathbf{t}$ over $\mathbf{y}$ for a new data point $\xpredin$ as
\begin{equation}
    p(\tpredc = 1|\xpredin) = \int_{\ypredc} p(\tpredc = 1|\ypredc)p(\ypredc|\xpredin)d\ypredc,
    \label{eq:dist_clas}
\end{equation}
where $\ypredc$ is the $c$-th column of the regression output of the model. Also, from Eq. \eqref{eq:reg_mod} we know that $p(\tpredc = 1|\ypredc) = \sigma(\ypredc)$, hence we rewrite Eq. \eqref{eq:dist_clas} as
\begin{equation}
\label{eq:true_post}
    p(\tpredc = 1|\xpredin) = \int_{\ypredc} \sigma( \ypredc)p(\ypredc|\xpredin)d\ypredc.
\end{equation}

However, as this last equation does not have an analytical solution, we will follow the idea presented in chapter 4 of \cite{bishop2006pattern}, where we can perform the following approximation
\begin{equation}
     \int_{\ypredc} \sigma( \ypredc)p(\ypredc|\xpredin)d\ypredc  \simeq \sigma\left(\frac{\langle \ypredc\rangle }{(1 + \frac{\pi}{8}\Sigma_{\ypredc})^{\unmed}}\right),   
\end{equation}
where $\langle \ypredc\rangle$ and $\Sigma_{\ypredc}$ represent the mean and variance of the regression posterior predictive distribution $p(\ypredc|\xpredin)$ and $\Phi(\cdot)$ is the probit function and $\lambda$ a scale parameter. Now, to find $p(\ypredc|\xpredin)$ we need to marginalise over $\zpred$
\begin{equation}
\label{eq:conv_final}
    p(\ypredc|\xpredin) = \int p(\ypredc|\zpred) p\p*{\zpred|\xpredin} d\zpred,
\end{equation}
where both $p(\ypredc|\zpred)$ and $p\p*{\zpred|\xpredin}$ follow a Gaussian distribution, which implies that the resulting distribution of the convolution defined in Eq. (\ref{eq:conv_final}) will be also Gaussian. Hence, we can approximate $p\p*{\zpred|\xpredin}$ as:
\begin{equation}
\begin{split}
    &p(\ypredc|\xpredin) 
    \varpropto \int 
    exp\bigg(-\frac{\psi}{2}\p*{\ypredc-\zpred\VcT}^2-\frac{\tau}{2}\p*{\zpred-\ang{\zpred}} \p*{\zpred-\ang{\zpred}}^\top\bigg) d \zpred\\
    &= exp\bigg(-\frac{1}{2}\p*{
    \psi \ypredc^2
    + \tau\ang{\zpred}\ang{\zpred}^\top
    }\bigg) \\
    &\int 
    exp\bigg(-\frac{1}{2}\p*{
    \zpred \p*{\tau \Ik+ \psi \VcT\Vc} \zpredT
    - 2\zpred \p*{\tau\ang{\zpred}^\top + \psi \VcT \ypredc}
    }\bigg) d\zpred,
\end{split}
\end{equation}
where by operating and subsequently applying the woodbury matrix identity, we get
\begin{align}
    p(\ypredc|\xpredin) \propto& 
    exp\left(-\frac{1}{2}\left( 
    \ypredc^2 \p*{\frac{1}{\psi} + \frac{\Vc\VcT}{\tau}}^{-1}
    - 2 \ypredc \Vc \p*{\frac{1}{\psi} + \frac{\Vc\VcT}{\tau}}^{-1} \ang{\zpred}^\top
    \right.\right. \nonumber\\
    \eqline \left.\left. 
    + \ang{\zpred}\p*{\frac{1}{\tau} \Ik + \frac{\Vc\VcT}{\psi}}^{-1}\ang{\zpred}^\top
    \right)\right) \nonumber\\
\end{align}

Therefore, identifying terms, we get that
\begin{align}
    \p*{\ypredc|\xpredin} \eqeq \N\p*{\ang{\ypredc}, \Sigma_{\ypredc}}
\end{align}
where
\begin{align}
    & \ang{\ypredc} = \sum_{\rm m}^M\left[\mathbf{H}(m,s^{\rm (m)})_{\rm *,:}\right] \\
    &\Sigma_{\ypredc} = \frac{1}{\langle\psi\rangle} + \frac{\Sigma_{\Vc}}{\langle\tau\rangle},
\end{align}
where $\mathbf{H}(m,s^{\rm (m)}) =  \Xm\langle\Wm\rangle^\top$ if $s^{\rm (m)}=0$ or $\mathbf{H}(m,s^{\rm (m)}) =  \Xm\Xsmt\langle\mathbf{A}^{\rm (m)}\rangle$ if $s^{\rm (m)}=1$.

\section{Experiments}
\label{sec:exp}

This section presents the application of the proposed model over two different real biomedical databases: \href{https://biofind.loni.usc.edu/}{BioFIND} \cite{biofind} and \href{https://adni.loni.usc.edu/}{ADNI} \cite{adni}.

\subsection{Baselines and experimental setup}
\label{sec:baselines}

We compare the performance of our model with various state-of-the-art models. These baselines are organized into two different groups. (1) single-view models, which concatenate all the data to analyze it as a whole, and (2) multi-view models, as the proposed model. In the first group, where we defined models that need to concatenate all data views to analyze them as a whole (single-view models) we include: (1) Linear Support Vector Machine with $\ell$1 penalty (LSVM-$\ell$1), (2) Random Forest (RF), and (3) Sparse Gaussian Process with Automatic Relevance Determination (SGP+ARD). On the second group, where we defined some multi-input models, we included: (4) Multiview Classification with Cohesion and Diversity (MCCD) \cite{tao2018multiview}, (5) Adaptive-Weighting Discriminative Regression (AWDR) for multi-view classification \cite{yang2019adaptive}, (6) Cross Partial Multi-View Networks (CPM-Nets) \cite{zhang2019cross}, (7) Boosting-based multiview learning algorithm (PB-MVBoost) \cite{goyal2019multiview}, (8) Multi-view Partial Least Squares (MV-PLS) \cite{mou2017multiview}, (9) \cite{mou2017multiview}, and (10) Multi-kernel Support Vector Machine with sparse regularisation (Lasso-MkSVM) \cite{huang2011identifying}.

Additionally, two important considerations must be addressed. First, the hyperparameters of each model have been validated within standard ranges (for general models) or within the ranges recommended by the authors of each respective paper. Second, in the case of MCCD and AWDR, as they create large $D\times D$ matrices, we included a data reduction step, specifically one case with a PCA and the other with a linear kernel.

Also, to adjust the hyperparameters of each model we defined a 5-fold and 10-fold nested CV procedure for Parkinson's and Alzheimer's databases, respectively. Thus, inner CV was used to define hyperparameters and the outer fold for evaluation. To assess the performance of the models we defined the following metrics: accuracy, balanced accuracy, precision, recall, F1-score, and AUC. The implementation of BALDUR is available \href{https://github.com/albello1/BALDUR.git}{here}.

\subsection{Experiments with BioFIND database}
\label{sec:parkinson}

\subsubsection{Dataset description}

The database (\href{https://biofind.loni.usc.edu/}{BioFIND}) consists of data from  96 healthy participants and 119 patients with moderate to advanced Parkinson's disease (PD) and the goal is to classify between healthy and PD. For this experiment, we will use 5 different data modalities: (1) 18 \textbf{neuropsychological} features that correspond to the standard Montreal Cognitive Assessment (MoCA) test; (2) 19 \textbf{sleep} features that characterize the quality and abnormalities while sleeping; (3) 2 \textbf{socio-economic} factors; (4) 8 \textbf{demographical} features; and (5) 841,685 \textbf{genetic} features that represent a full genotyping of the most common human single nucleotide polymorphisms (SNPs).

\subsubsection{Performance Evaluation}
\label{sec:performance_park}

Table \ref{tab:Park_res} presents the performance of baseline models and BALDUR, demonstrating BALDUR’s superiority in accuracy, balanced accuracy, and AUC. Excluding models that classify all predictions into a single category, BALDUR achieved the highest recall, indicating its robustness despite the high-dimensional and sparse nature of the genetic view, which limits generalizability and leads many models to perform random classification. BALDUR and Lasso-MkSVM, a close competitor, proved effective in handling such complexity; however, BALDUR excelled by jointly performing FS and latent variable inference in an iterative loop, enabling it to filter out irrelevant/redundant features when generating the latent space in a more efficient way than Lasso-MkSVM. BALDUR achieved notable compactness in FS, selecting the most reduced feature subset ($9.5 \times 10^{-4} \%$), unlike other models that retained irrelevant features across views (or the whole feature set), resulting in poorer diagnostic accuracies.

 \begin{table*}[ht!]
 \centering
 \vspace{0.4cm}
\begin{adjustbox}{max width=1\textwidth}
 \begin{tabular}{c c c c c c c |c}
 \hline
    &   Accuracy    &  bal Accuracy    &   Precision   &  Recall    &   F1   & AUC & $\%$ Features selected \\ \hline
    LSVM-$\ell$1   &    0.51 $\pm$ 0.02   & 0.51 $\pm$ 0.02  & 0.58 $\pm$ 0.02   &  0.54 $\pm$ 0.03  &  0.56 $\pm$  0.02 & 0.48 $\pm$ 0.05 & 6.78 $\pm$ 0.56 $\%$ \\
    RF   &   0.56 $\pm$ 0.01   &  0.49   $\pm$ 0.01  & 0.57 $\pm$  0.01    &  0.95 $\pm$  0.03$^{*}$   &  0.71 $\pm$ 0.01    & - & 45.56 $\pm$ 34.03 $\%$ \\ 
    SGP+ARD   & 0.57   $\pm$ 0.01  &   0.50  $\pm$ 0.00  &  0.57 $\pm$ 0.01  &   1.00 $\pm$  0.00$^{*}$   &  0.73 $\pm$  0.01   & 0.58 $\pm$ 0.06 & 100$\%$  \\
    PCA-MCCD   &    0.57 $\pm$ 0.01    &   0.50  $\pm$ 0.0  & 0.57  $\pm$ 0.01   &  1.00 $\pm$  0.00$^{*}$   & 0.73  $\pm$  0.01   &  - & 100$\%$  \\
    K-MCCD   &     0.57 $\pm$ 0.01    &   0.50  $\pm$ 0.0  & 0.57  $\pm$ 0.01   &  1.00 $\pm$  0.00$^{*}$   & 0.73  $\pm$  0.01   &  - & 100$\%$  \\
    PCA-AWDR   &    0.56 $\pm$ 0.02   &   0.49  $\pm$  0.01 &  0.56  $\pm$  0.04 &  0.93 $\pm$ 0.13$^{*}$    &  0.70 $\pm$ 0.06    &  - & 100$\%$  \\
    K-AWDR   &     0.56 $\pm$ 0.02   &   0.49  $\pm$  0.01 &  0.56  $\pm$  0.04 &  0.93 $\pm$ 0.13$^{*}$    &  0.70 $\pm$ 0.06    &  - & 100$\%$  \\ 
    CPM-Nets   &     0.57 $\pm$ 0.01    &   0.50  $\pm$ 0.0  & 0.57  $\pm$ 0.01   &  1.00 $\pm$  0.00$^{*}$   & 0.73  $\pm$  0.01   &  - & 100$\%$  \\
    PB-MVBoost   &     0.57 $\pm$ 0.01    &   0.50  $\pm$ 0.0  & 0.57  $\pm$ 0.01   &  1.00 $\pm$  0.00$^{*}$   & 0.73  $\pm$  0.01   &  - & 100$\%$  \\
    MV-PLS   &    0.43 $\pm$ 0.02   &   0.49  $\pm$  0.05 & 0.52  $\pm$ 0.12  & 0.38 $\pm$  0.22   &  0.41 $\pm$  0.20   & 0.49 $\pm$ 0.06 & 100 $\%$ \\
    Lasso-MkSVM   &    0.66 $\pm$ 0.07   &  0.66  $\pm$ 0.07  & \bf 0.64  $\pm$ 0.13  & 0.73 $\pm$  0.06   & \bf 0.68  $\pm$ 0.10    &  0.68 $\pm$ 0.08 & 5.68 $\pm$ 0.98 $\%$\\  
    BALDUR   &   \bf 0.68 $\pm$ 0.05   &  \bf 0.70  $\pm$ 0.05  &  0.55  $\pm$ 0.08 & \bf 0.82 $\pm$  0.09   & 0.66  $\pm$  0.08   & \bf 0.73 $\pm$ 0.06 & \footnotesize $\boldsymbol{10^{-4}\cdot(9.50}\pm \boldsymbol{0.76)} \%$ \\ \hline
 \end{tabular}
 \end{adjustbox}
 \caption{Performance results on the Parkinson's database. The rows represent the models under study and the columns the different metrics; the final column includes the average ($\%$) of final features selected by the model. For models that do not generate soft labels when predicting, some AUC spaces have been left blank. (\textbf{*}) Denotes a chance level classification performance.}
 \label{tab:Park_res}
 \end{table*}

\subsubsection{Biomarker Selection}

\begin{figure*}[ht!]
    \centering
    \includegraphics[width=\linewidth]{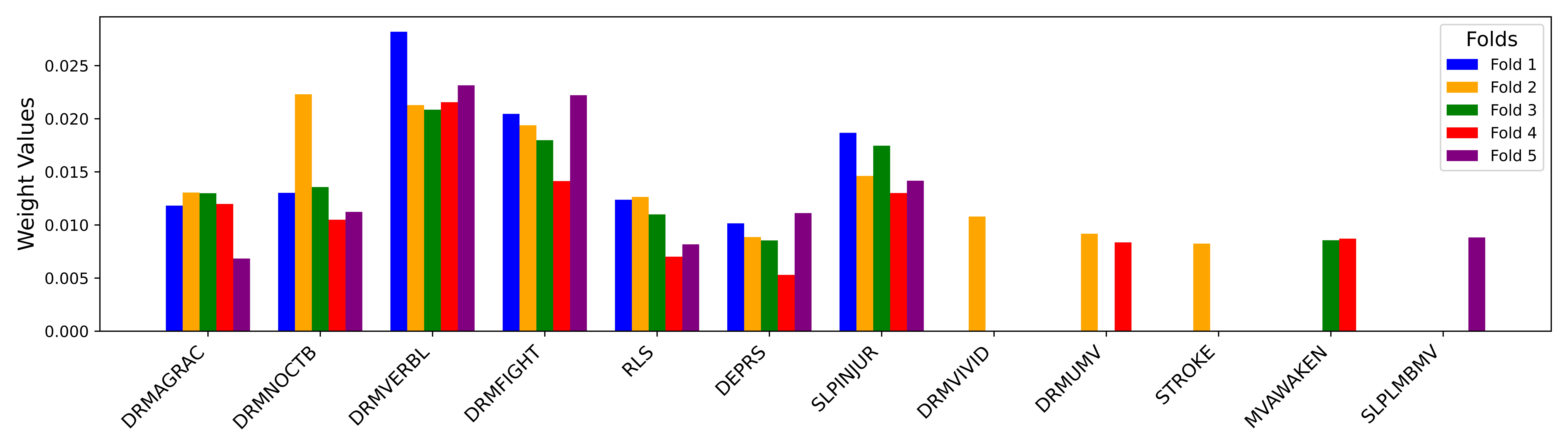}
    \caption{Features selected per fold (x-axis) and their corresponding weight in absolute value (y-axis).}
  \label{fig:histogram} 
\end{figure*}

Figure \ref{fig:histogram} depicts the features selected per fold and their corresponding weight in absolute value. The model generated a single latent vector that exclusively projects the view containing the sleep data modality, while the remaining modalities were removed through the pruning process. In all folds, 7 common features have been selected (green bars): DRMAGRAC, DRMNOCTB, DRMVERBL, DRMFIGH, RLS, SLPINJUR, and DEPRS. These features correspond to: Aggressive or action-packed dreams (DRMAGRAC), Noctural behavior (DRMNOCTB), Speaking sleep (DRMVERBL), Sudden limb movements (DRMFIGH), Restless Leg Syndrome (RLS), Hurt bed partner (SLPINJUR), and Depression (DEPRS). A review of the scientific literature reveals several studies linking sleep disorders to the onset of Parkinson’s disease. Some recent studies, such as \cite{thangaleela2023neurological} and \cite{dodet2024sleep}, identify general sleep patterns in the Parkinson's population, especially the ones related to insomnia, such as nocturnal behavior and, mainly, RLS \cite{maggi2024prevalence,diaconu2023restless}. We found some studies that link abnormal (aggressive or stressful) dreams (DRMAGRAC) to the disorder \cite{diederich2008visual,otaiku2022distressing} and others that relate depression (DEPRS) too \cite{lieberman2006depression,starkstein1990depression}. However, despite being the most relevant feature in 4 out of 5 folds, no specific study related to speaking during sleep (DRMVERBL) has been found.

\subsection{Experiments with ADNI database}

\subsubsection{Dataset description}

We compare the performance of the algorithms in magnetic resonance imaging (MRI)-based classification between early and late mild cognitive impairment (MCI) using the ADNI dataset \cite{adni}. ADNI separates these stages based on the same single episodic memory measure that is used to diagnose MCI (i.e., one story from the Wechsler Memory Scale-Revised [WMS-R] Logical Memory II subtest) \cite{edmonds2019early}. However, whether these two stages can be separated based on MRI data only is an interesting question.

Data used in this work were obtained from the ADNI \url{http://adni.loni.usc.edu}. The ADNI was launched in 2003 as a public-private partnership, led by Principal Investigator Michael W. Weiner, MD. For up-to-date information, see \url{http://www.adni-info.org.} We included T1-weighted (T1w) MRI data from 661 participants with late MCI (LMCI) and 402 participants with early MCI (EMCI).  Each sample contained 5 different views of participants' MRI scan: (1) T1w intensity images registered to the stereotactic space and resampled to 4$mm^3$ voxels (37 643 features); (2) Gray matter density images registered to the stereotactic space and resampled to 4$mm^3$ voxels; (3 -5) gray matter, white matter, and cerebrospinal fluid volumes in 136 ROIs defined by the neuromorphometrics atlas. All the image pre-processing was performed using CAT12 tool (Version 8.1) \cite{gaser2024cat}.

\subsubsection{Performance Evaluation}
\label{sec:perf_alzh}

Table \ref{tab:per_alz} shows that BALDUR outperformed all baselines on the ADNI dataset, achieving superior accuracy, balanced accuracy, recall, F1 score, and AUC. Although PCA-AWDR matched BALDUR in accuracy, BALDUR’s higher balanced accuracy (0.80 vs. 0.70) indicates most balanced classification. BALDUR also surpassed SGP+ARD, which, despite high precision, has lower recall and overfits the majority class, whereas BALDUR maintained a balanced classification, showing a 4$\%$ improvement in AUC and F1 scores. Additionally, BALDUR excelled in feature selection, achieving the best performance among the models with only 2.36$\%$ of features, corresponding to a final single view, avoiding the redundancy issues seen in Lasso-MkSVM and LSVM-$\ell$1, which over-regularize and lead to poor generalization when including information from all views.

\begin{table*}[ht!]
 \centering
 \vspace{0.2cm}
\begin{adjustbox}{max width=\textwidth}
 \begin{tabular}{c c c c c c c |c}
 \hline
    &   Accuracy    &  bal Accuracy    &   Precision   &  Recall    &   F1   & AUC & $\%$ Features selected\\ \hline
    LSVM-$\ell$1   &  0.68  $\pm$  0.06  & 0.70 $\pm$  0.06 & \bf 0.82 $\pm$ 0.05   &  0.62 $\pm$ 0.09  &  0.70 $\pm$ 0.07  &  0.75 $\pm$ 0.05 & $\boldsymbol{0.23 \% \pm 0.03}$\\
    RF   & 0.75  $\pm$  0.03  & 0.74 $\pm$ 0.03  & 0.81 $\pm$  0.02  &  0.80 $\pm$ 0.04  & \bf 0.80 $\pm$ 0.03  & - & 75.64 $\%$ $\pm$ 18.16 \\ 
    SGP+ARD   &  0.74 $\pm$ 0.03  &   0.75  $\pm$  0.02 & \bf 0.87 $\pm$ 0.02  & 0.67 $\pm$ 0.05    &  0.76 $\pm$  0.03   & 0.81 $\pm$ 0.04 & 100$\%$  \\
    PCA-MCCD   &  0.62  $\pm$  0.001  & 0.5 $\pm$ 0.0  & 0.62 $\pm$ 0.01   &  1.00  $\pm$ 0.00$^{*}$  & 0.76  $\pm$ 0.01  &  - & 100$\%$  \\
    K-MCCD   &    0.62  $\pm$  0.001  & 0.5 $\pm$ 0.0  & 0.62 $\pm$ 0.01   &  1.00  $\pm$ 0.00$^{*}$  & 0.76  $\pm$ 0.01  &  - & 100$\%$  \\
    PCA-AWDR   & \bf 0.78  $\pm$ 0.04   & 0.70 $\pm$ 0.02 & 0.78 $\pm$ 0.04   & 0.77  $\pm$ 0.02  & 0.78  $\pm$ 0.02   &  - & 100$\%$ \\
    K-AWDR   &  0.77   $\pm$  0.02  & 0.75 $\pm$ 0.03  & 0.80 $\pm$  0.02  &  0.82 $\pm$ 0.02  & \bf 0.80 $\pm$ 0.02  &  - & 100$\%$ \\
    CPM-Nets   &   0.62  $\pm$  0.001  & 0.5 $\pm$ 0.0  & 0.62 $\pm$ 0.01   &  1.00  $\pm$ 0.00$^{*}$  & 0.76  $\pm$ 0.01  &  - & 100$\%$ \\
    PB-MVBoost   &   0.61  $\pm$  0.01  & 0.48 $\pm$  0.01 & 0.61 $\pm$ 0.01   &   0.97 $\pm$ 0.01$^{*}$  & 0.75  $\pm$  0.01 &  - & 100$\%$ \\
    MV-PLS   &  0.67  $\pm$ 0.02   & 0.70 $\pm$  0.04 & 0.80 $\pm$    0.05& 0.63  $\pm$ 0.05  & 0.70 $\pm$ 0.04  & 0.68 $\pm$ 0.06 & 100$\%$\\
    Lasso-MkSVM   &   0.68 $\pm$ 0.06   & 0.61 $\pm$ 0.05  & 0.72 $\pm$ 0.10   &  0.75 $\pm$ 0.04  &  0.73 $\pm$ 0.06  & 0.63 $\pm$ 0.07 & 0.54 $\%$ $\pm$ 0.11\\
    BALDUR   &  \bf 0.78 $\pm$  0.03  & \bf 0.80 $\pm$ 0.02  & 0.80 $\pm$ 0.12   &\bf  0.93 $\pm$ 0.02  & \bf 0.80 $\pm$  0.03 & \bf 0.85 $\pm$ 0.03 & 2.36 $\%$ $\pm$ 0.98\\\hline
 \end{tabular}
 \end{adjustbox}
 \caption{Performance results on the ADNI database. The rows represent the models under study and the columns the different metrics; the final column includes the average ($\%$) of final features selected by the model. For models that do not generate soft labels when predicting, some AUC spaces have been left blank. (\textbf{*}) Denotes a chance level classification performance.
 %It follows the same structure as Table \ref{tab:Park_res}.
 }
 \label{tab:per_alz}
 \end{table*}

\subsubsection{Biomarker Selection}
\label{sec:biomarker_alz}

\begin{figure}[ht!]
    \centering
    % First row
    \begin{subfigure}[b]{0.278\columnwidth}
        \centering
        \includegraphics[width=\linewidth]{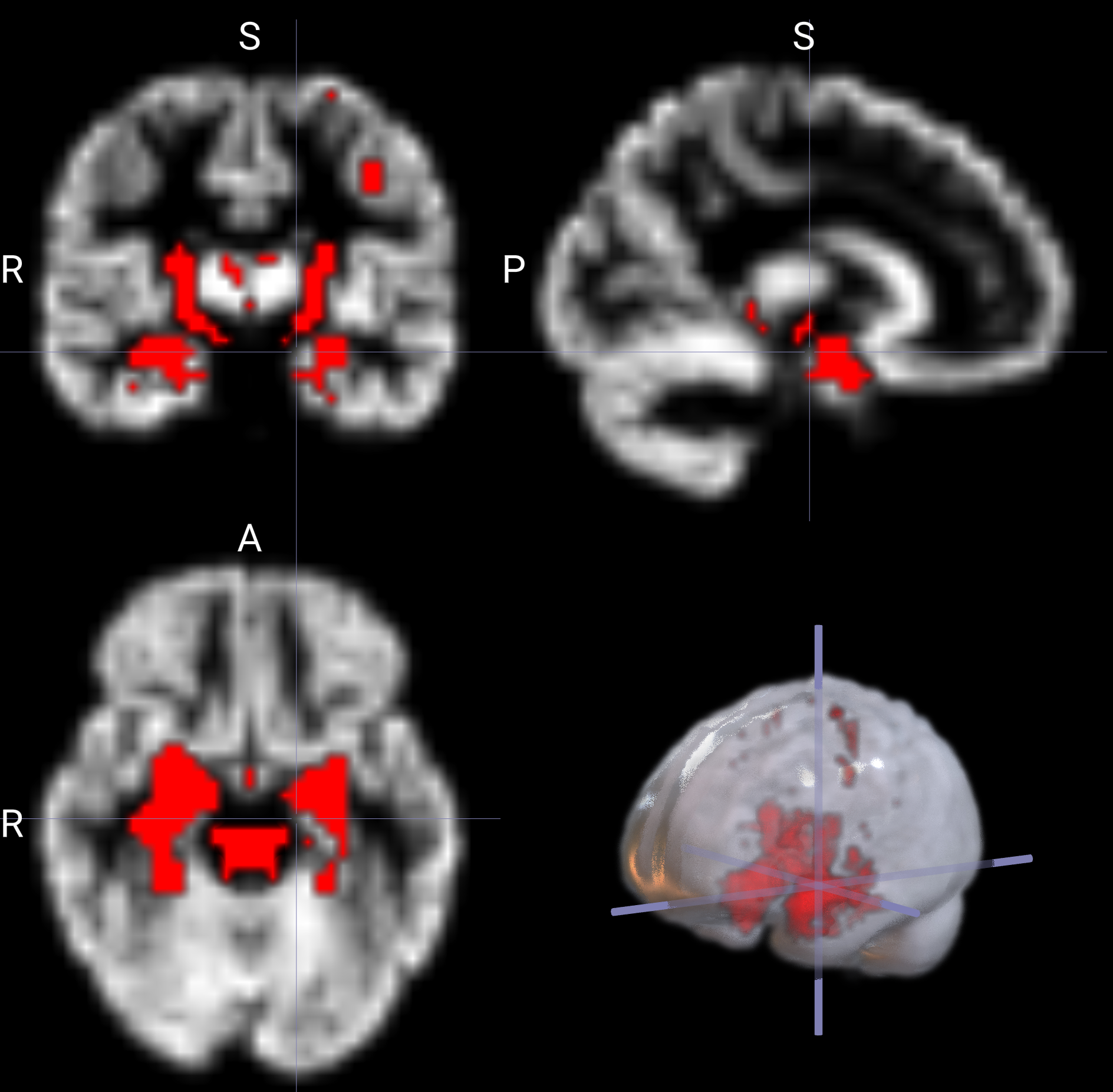}
        \caption{Gray matter}
    \end{subfigure}
    % \vskip\baselineskip
    \begin{subfigure}[b]{0.27\columnwidth}
        \centering
        \includegraphics[width=\linewidth]{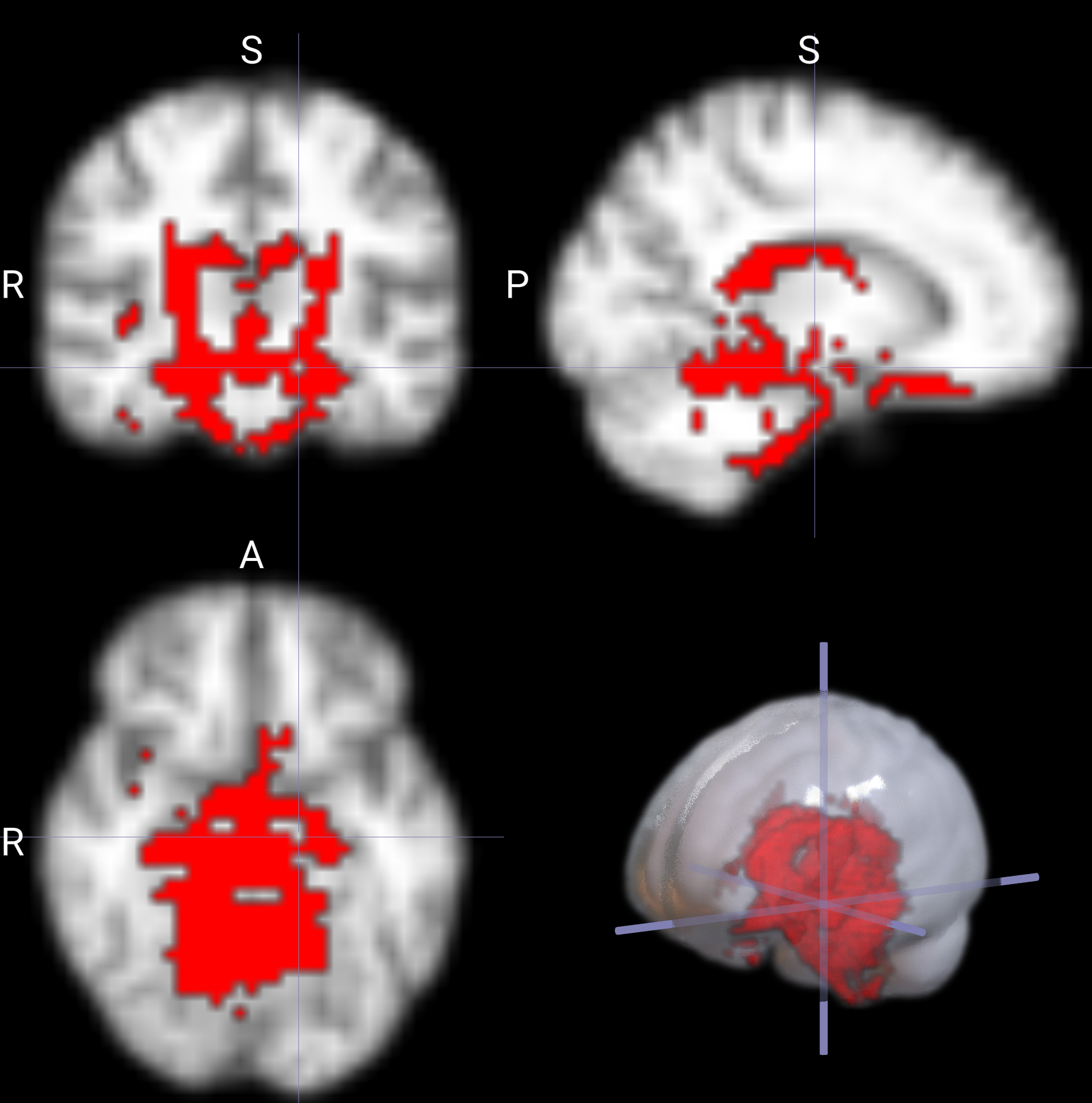}
        \caption{Intensity}
    \end{subfigure}

    \begin{subfigure}[b]{0.27\columnwidth}
        \centering
        \includegraphics[width=\linewidth]{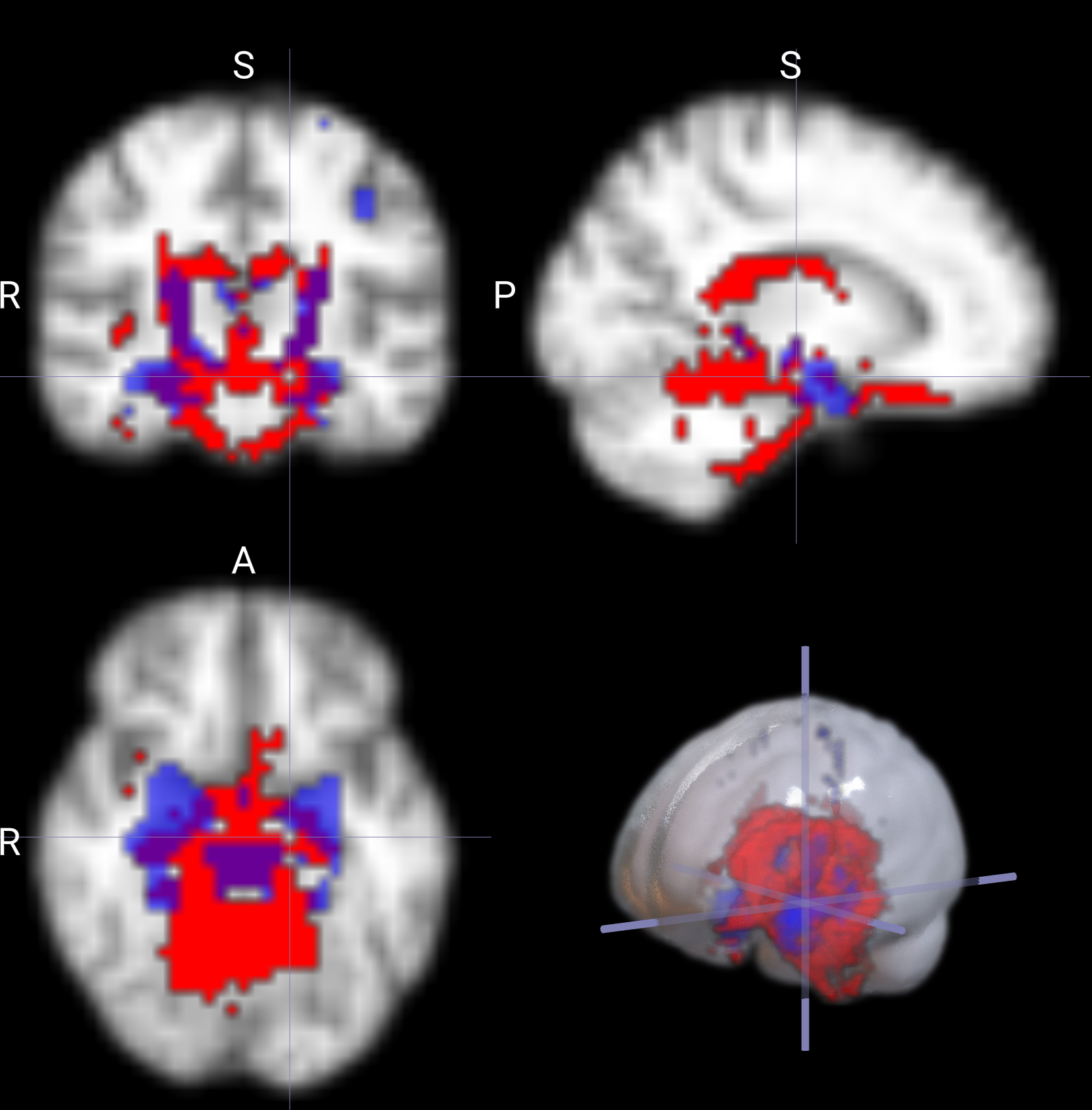}
        \caption{Both}
    \end{subfigure}

% MNI coordinates - 16 mm, -16 mm, -16 mm
    \caption{Brain regions selected by BALDUR at gray matter (top left) and intensity (top right). Also, the bottom image depicts the overlap between gray matter (blue) and intensity (red).}
    \label{fig:Nueoimage}
\end{figure}
The model yielded two distinct behaviors across the 10 CV folds: it utilized either gray matter density maps or intensity images (but not both) and consistently removed the ROI views. This was expected as gray matter density and intensity images contained overlapping information. Interestingly, the ROI views that can be seen as a summarization of the whole-image features were consistently removed. Figure \ref{fig:Nueoimage} depicts the final brain regions selected over the gray matter (a), intensity images (b), and the overlap between both (c). For gray matter density map, BALDUR highlighted a large cluster extending over bilateral hippocampus, amygdala, and parahippocampal cortex, which are central to the memory function, and smaller clusters in thalamus, cerebellum, and precentral cortex. Thalamus and cerebellum have been linked to memory function \cite{fama2015thalamic,zhang2023cerebellum}. For intensity map, BALDUR highlighted a large cluster containing ventricles but extending to hippocampus, cerebellum, and thalamus as well as a smaller cluster in insula. Ventricles can be expected to be enlarged in later stages of MCI \cite{apostolova2012hippocampal} and insula has been linked with dementias, including Alzheimer's disease \cite{liu2018altered}.

\section{Conclusions}
\label{sec:Conclusions}

This paper introduces the BALDUR model, a Bayesian algorithm designed to efficiently deal with multi-modal biomedical datasets that encompass high-dimensional data views and reduced sample sizes, addressing this challenge through an efficient view combination over a common latent space. Also, it can operate in dual or primal spaces, enabling it to handle both wide and normal data views simultaneously. As demonstrated in both BioFIND and ADNI databases, the proposed model stands out in comparison with the benchmarks when providing generalizable solutions. That is, the model enhances its performance and mitigates overfitting risks by means of an iterative FS scheme that systematically removes irrelevant or redundant information, including entire data views when necessary, to avoid adding noise to the latent space generation. Finally, its linear structure further allows to generate explainable solutions, enhancing the model's decision-making understanding and allowing for the description of statistical surrogate markers.
\\

% \noindent\textbf{CRediT authorship contribution statement}
% \\

% \textbf{Albert Belenguer-Llorens: }Conceptaulization, Data curation, Methodology, Investigation, Validation, Software,Writing - original draft. \textbf{Carlos Sevilla-Salcedo: }Conceptualization, Methodology, Investigation, Software,Writing - review $\&$ editing \textbf{Emilio Parrado-Hernandez: }Conceptualization, Funding acquisition, Software, Supervision, Writing - review $\&$ editing \textbf{Vanessa Gomez-Verdejo: }Conceptualization, Investigation, Funding acquisition, Project administration, Supervision, Writing - review $\&$ editing. All authors have read and agreed to the published version of the manuscript.
% \\

% \noindent\textbf{Declaration of competing interest }
% \\

% The authors declare that they have no known competing financial interests or personal relationships that could have appeared to influence the work reported in this paper. 
% \\

% \noindent\textbf{Data availability }
% \\

% The authors do not have permission to share data. 
% \\

% \noindent\textbf{Acknowledgements}
% \\

% This work is partially supported by grants PID2020-115363RB-I00 funded by MCIN/AEI/10.13039/50110001103, Spain. V.G-V and C.S-S's work is partially supported by TED2021-132366B-I00 funded by MCIN/AEI/10.13039/501100011033 and by the "European Union NextGenerationEU/PRTR".

\newpage
\appendix
\setcounter{equation}{0}
\setcounter{figure}{0}
\setcounter{table}{0}
\setcounter{section}{0}
\makeatletter
\renewcommand{\theequation}{A.\arabic{equation}}
\renewcommand{\thefigure}{A\arabic{figure}}

\section{Logistic Regression Bound}
\label{sec:Ap_logbound}

As we work in a binary framework, the probability of the logistic regression $p(\mathbf{t}_{\rm n,c}|y_{\rm n,c}) = e^{y_{\rm n,c} t_{\rm n,c}}\sigma(-y_{\rm n,c})$ can be expressed as
\begin{equation}
    p(t_{\rm n,c} = 1|y_{\rm n,c}) = e^{\yn}\sigma(-y_{\rm n,c}) = %\frac{e^{\yn}e^{-\yn}}{1 + e^{-\yn}} = 
    (1+e^{-\yn})^{-1}.
\end{equation}

Note that we used a formulation $\mathbf{Y} \in \R^{N,C}$

Also, we can express the log-likelihood of $p(t_{\rm n,c} = 1|y_{\rm n,c})$ as
\begin{equation}
\label{eq:E2}
    \ln p(t_{\rm n,c} = 1|y_{\rm n,c}) = - \ln(1+e^{-y_{\rm n,c}}) = \frac{y_{\rm n,c}}{2} - \ln(e^{y_{\rm n,c}}+e^{-y_{\rm n,c}}),
\end{equation}
where $f(y_{\rm n,c})=-\ln(e^{y_{\rm n,c}}+e^{-y_{\rm n,c}})$ is a convex function in $y_{\rm n,c}^2$. As demonstrated at \cite{jaakkola2000bayesian}, we can use a curve to lower bound a convex function. Hence, we lower bound $f(y_{\rm n,c})$ with the first-order Taylor series expansion in $y_{\rm n,c}^2$ as
\begin{equation}
\begin{split}
\label{eq:E3}
    f(y_{\rm n,c}) & \geq f(\xi_{\rm n,c}) + \frac{\partial f(\xi_{\rm n,c})}{\partial \xi_{\rm n,c}^2}(y_{\rm n,c}^2 - \xi_{\rm n,c}^2) = -\frac{\xi_{\rm n,c}}{2} + \ln p(t_{\rm n,c} = 1|\xi_{\rm n,c}) \\ & + \frac{1}{4\xi_{\rm n,c}} tgh\left(\frac{\xi}{2}\right)(y_{\rm n,c}^2 - \xi_{\rm n,c}^2),
\end{split}
\end{equation}
where $\xi_{\rm n,c}^2$ is the center of the Taylor serie and $f(y_{\rm n,c}) = f(\xi_{\rm n,c}) + \frac{\partial f(\xi_{\rm n,c})}{\partial \xi_{\rm n,c}^2}(\yn^2 - \xi_{\rm n,c}^2)$ when $\xi_{\rm n,c}^2 = y_{\rm n,c}^2$. Now, as presented in \cite{jaakkola2000bayesian}, if we combine Eq. (\ref{eq:E2}) and Eq. (\ref{eq:E3}) we achieve the lower bound over $p(t_{\rm n,c} = 1|y_{\rm n,c})$:
\begin{equation}
    p(t_{\rm n,c} = 1|y_{\rm n,c}) = 
    e^{y_{\rm n,c}}\sigma(-y_{\rm n,c}) \geqslant h(y_{\rm n,c}, \xi_{\rm n,c}) = 
    e^{y_{\rm n,c} t_{\rm n,c}}\sigma(\xi_{\rm n,c})e^{-\frac{y_{\rm n,c} + \xi_{\rm n,c}}{2} - \lambda(\xi_{\rm n,c})(y_{\rm n,c}^2 - \xi_{\rm n,c}^2)},
\end{equation}
where $\lambda(a) = \frac{1}{2a}(\sigma(a) - \unmed)$.
\newpage
\appendix
\setcounter{equation}{0}
\setcounter{figure}{0}
\setcounter{table}{0}
\setcounter{section}{1}
\makeatletter
\renewcommand{\theequation}{B.\arabic{equation}}
\renewcommand{\thefigure}{B\arabic{figure}}

\section{Variational Inference of BALDUR}
\label{sec:Ap_inf_BALDUR}

This section presents the mathematical developments of the approximated posterior distributions of the model variables, i.e., $q(\mathbf{\Theta})$. That is,
\begin{equation}
\begin{split}
    p(\boldsymbol{\Theta}| \mathbf{X}^{\M}, \mathbf{t})  \backsimeq   q(\boldsymbol{\Theta}) = & \prod_{m}^{M}\left(q\left(\Wm\right)^{\mathbbm{1}(s^{\rm (m)} = 0)}q\left(\mathbf{A}^{\rm (m)}\right)^{\mathbbm{1}(s^{\rm (m)} = 1)}\prod_{\rm k}^{K}q\left(\delta_{\rm k}\right)\prod_{\rm d}^{D}q\left(\gamma_{\rm d}\right)\right) \\ &\prod_n^{N}\left(q\left(\mathbf{z}_{\rm n,:}\right)q\left(\mathbf{y}_{\rm n,:}\right)\right)q(\mathbf{V})\prod_k^K\left(q(\omega_{\rm k})\right)q(\tau)q(\psi).
\end{split}
\end{equation}

Also, as presented in Section \ref{sec:generative}, the model variables follow these prior distributions:
\begin{align}
    \mathbf{z}_{\rm n,:} \eqsimil  \N\p*{\sum_{\rm m}^M\mathbf{P}(m,s^{\rm (m)})_{n,:}\p*{\tau}^{-1}\Ik} \hspace{1.1cm} {\rm n} = 1,\ldots, N \label{eq:zprior_BALDUR}\\
    \Wkd  \eqsimil \N\p*{0,\p*{\delta_{\rm k}^{\rm (m)} \gamd}^{-1}} \hspace{2.7cm} {\rm k} = 1,\ldots, K \hspace{0.1cm} \text{and} \hspace{0.1cm} {\rm d} = 1,\ldots, D^{(m)} \label{eq:Wprior_sparse} \\
    \delta_{\rm k}^{\rm (m)} \eqsimil \Gamma\p*{\alpha^{\delta^{\rm (m)}}_{\rm k}, \beta^{\delta^{\rm (m)}}_{\rm k}} \hspace{3.85cm} {\rm k} = 1,\ldots, K \label{eq:aprior}\\
    \gamd \eqsimil \Gamma\p*{\alpha^{\gamma^{\rm (m)}}_{\rm d}, \beta^{\gamma^{\rm (m)}}_{\rm d}} \hspace{3.79cm} {\rm d} = 1,\ldots, D^{(m)} \label{eq:gprior_sparse}\\
    \tau \eqsimil \Gamma\p*{ \alpha^{\tau}, \beta^{\tau} } \label{eq:tprior}\\
    \mathbf{y}_{\rm n,:} \eqsimil  \N\p*{\Zn \VT,\p*{\psi}^{-1}\Ic}\\
    \Vk  \eqsimil \N\p*{0,(\omega_{\rm k}^{-1})\Ic} \hspace{3.85cm} {\rm n} = 1,\ldots, N\label{eq:Vprior_sparse} \\
    \omega_{\rm k} \eqsimil \Gamma\p*{\alpha^{\omega}_{\rm k}, \beta^{\omega}_{\rm k}} \hspace{4.8cm} {\rm k} = 1,\ldots, K \label{eq:bprior}\\
    \psi \eqsimil \Gamma\p*{ \alpha^{\psi}, \beta^{\psi}}, \label{eq:pprior} 
\end{align}
where
\begin{equation}
\label{eq:P_simp}
    \mathbf{P}(m,s^{\rm (m)})_{n,:} = \Bigg\{ \begin{array}{rcl}
        \mathbf{x}_{n,:}\Wm^\top & \mbox{if}
        &s^{\rm (m)}=0 \\ \mathbf{x}_{n,:}\Xsmt\mathbf{A}^{\rm (m)} & \mbox{if} &s^{\rm (m)}=1
        \end{array},
\end{equation}

Furthermore, to clarify the relationships between the model variables $\mathbf{\Theta}$, and apply Eq. \eqref{eq:lnqopt}, we define the joint model distribution as:
\begin{equation}
\begin{split}
    p(\mathbf{Y},\mathbf{Z},\tau, \psi, \boldsymbol{\omega},\mathbf{V}, \boldsymbol{\delta^{\M}}, \boldsymbol{\gamma^{\M}}, \mathbf{W}^{\M}\mathbbm{1}(s^{\M}=0), \mathbf{A}^{\M}\mathbbm{1}(s^{\M}=1)) = p(\mathbf{T}|\mathbf{Y})p(\mathbf{Y}|\mathbf{Z}, \mathbf{V},\psi) \\
    p(\mathbf{V}|\boldsymbol{\omega})p(\boldsymbol{\omega})p(\boldsymbol{\psi})p(\mathbf{Z}|\mathbf{W}^{\M}\mathbbm{1}(s^\M = 0),\mathbf{A}^{\M}\mathbbm{1}(s^\M = 1),\tau)p(\mathbf{W}^{\M}|\boldsymbol{\gamma}^\M, \boldsymbol{\delta}^\M)^{\mathbbm{1}(s^\M = 0)} \\
    p(\mathbf{\tilde{X}}^{\M\top}\mathbf{A}^{\M}|\boldsymbol{\gamma}^\M, \boldsymbol{\delta}^\M)^{\mathbbm{1}(s^\M = 1)} p(\boldsymbol{\gamma}^\M) p(\boldsymbol{\delta}^\M) p(\tau)
\end{split}
\end{equation}

\subsection{Mean Field Approximation of $\mathbf{Z}$}

Following the mean-field procedure defined in Eq. \eqref{eq:lnqopt}, we define the approximated posterior distribution of $\mathbf{Z}$ as:
\begin{equation}
\label{eq:e_z}
    \ln q(\mathbf{Z}) = \E[\ln p(\mathbf{Y}|\mathbf{Z}, \mathbf{V},\psi) + \ln p(\mathbf{Z}|\mathbf{W}^{\M}\mathbbm{1}(s^\M = 0),\mathbf{A}^{\M}\mathbbm{1}(s^\M = 1),\tau)].
\end{equation}

Thus, we can develop the first term of the summation as:
\begin{equation}
\begin{split}
    \ln p(\mathbf{Y}|\mathbf{Z}, \mathbf{V},\psi) & = \sum_{n=1}^N \ln \N(\mathbf{z}_{n,:}\mathbf{V}^\top,\psi) \\
    & = \sum_{n=1}^{N}\left(-\unmed\lnpi + \unmed\ln \psi -\frac{\psi}{2}(\yn - \mathbf{z}_{n,:}\mathbf{V}^\top)(\Yn - \mathbf{z}_{n,:}\mathbf{V}^\top)^\top\right) \\
    & = \sum_{n=1}^{N}\left(-\unmed\lnpi + \unmed\ln \psi -\frac{\psi}{2}\left(\Yn\YnT -2\Zn\VT\YnT + + \Zn\VT\V\ZnT\right)\right)
    \label{eq:yz}
\end{split}
\end{equation}
as we will calculate the expectation over $\Z$, we can rewrite Eq. (\ref{eq:yz}) as
\begin{equation}
\label{eq:term1_z}
    \ln p(\mathbf{Y}|\mathbf{Z}, \mathbf{V},\psi) = \sum_{n=1}^N\left(-\unmed\psi\left(-2\Zn\VT\YnT + \Zn\VT\V\ZnT\right)\right) + \const.
\end{equation}

Now, the second term of the summation can be expressed as:
\begin{equation}
\begin{split}
\label{eq:zy}
    & \ln p(\mathbf{Z}|\mathbf{W}^{\M}\mathbbm{1}(s^\M = 0),\mathbf{A}^{\M}\mathbbm{1}(s^\M = 1),\tau) \\
    & = \sumn\ln\N\left(\summ\left(\left(\mathbf{x}_{n,:}\Wm^\top\right)\mathbbm{1}(s^{(m)} = 0) + \left(\mathbf{x}_{n,:}\Xsmt\mathbf{A}^{\rm (m)}\right)\mathbbm{1}(s^{(m)} = 1)\right), \tau\right)
\end{split}
\end{equation}
also, using Eq. (\ref{eq:P_simp}), we can simplify Eq. (\ref{eq:zy}) as:
\begin{equation}
\begin{split}
\label{eq:final_z}
    & \ln p(\mathbf{Z}|\mathbf{W}^{\M}\mathbbm{1}(s^\M = 0),\mathbf{A}^{\M}\mathbbm{1}(s^\M = 1),\tau) \\
    & = \sumn \Bigg(-\unmed\lnpi + \unmed\ln \tau - \frac{\tau}{2}\left(\Zn- \summ \mathbf{P}(m,s^{\rm (m)})_{n,:}\right) \left(\Zn - \summ \mathbf{P}(m,s^{\rm (m)})_{n,:}\right)^\top\Bigg).
\end{split}
\end{equation}

As we will calculate the expectation over $\Z$, we can set as constant values the elements that are not multiplying $\Z$. Thus, we simplify Eq. (\ref{eq:final_z}) as:
\begin{equation}
\label{eq:term2_z}
\begin{split}
    &\ln p(\mathbf{Z}|\mathbf{W}^{\M}\mathbbm{1}(s^\M = 0),\mathbf{A}^{\M}\mathbbm{1}(s^\M = 1),\tau) \\ & = \sumn \left(-\unmed\tau\left(\Zn\ZnT - \Zn\summ \mathbf{P}(m,s^{\rm (m)})_{n,:}^\top\right)\right) + \const.
\end{split}
\end{equation}

Hence, if we merge fill Eq. (\ref{eq:e_z}) with both Eq. (\ref{eq:term1_z}) and (\ref{eq:term2_z}), we obtain:
\begin{equation}
\begin{split}
    &\ln q(\mathbf{Z}) = \E\Bigg[-\sumn\Bigg(\unmed\psi\left(-2\Zn\VT\YnT + \Zn\VT\V\ZnT\right)\\ & +\unmed\tau\left(\Zn\ZnT - \Zn\summ \mathbf{P}(m,s^{\rm (m)})_{n,:}^\top\right)\Bigg)\Bigg],
\end{split}
\end{equation}
where, if we apply the expectation and reorganize the elements, we obtain
\begin{equation}
\begin{split}
    \ln q(\mathbf{Z}) = \sumn\Bigg(-\Zn\left(\unmed\langle\psi\rangle\langle\VT\V\rangle + \unmed\langle\tau\rangle\right)\ZnT + \Zn\left(2\langle\VT\rangle\langle\YnT\rangle + \summ \mathbf{H}(m,s^{\rm (m)})_{n,:}^\top\right)\Bigg),
\end{split}
\end{equation}
where 
\begin{equation}
\label{eq:Hh}
    \mathbf{H}(m,s^{\rm (m)})_{n,:} = \Bigg\{ \begin{array}{rcl}
        \mathbf{x}_{n,:}\langle\Wm^\top\rangle & \mbox{if}
        &s^{\rm (m)}=0 \\ \mathbf{x}_{n,:}\Xsmt\langle\mathbf{A}^{\rm (m)}\rangle & \mbox{if} &s^{\rm (m)}=1
        \end{array},
\end{equation}

Thus, we can identify the different elements and subsequently parameterize the approximated posterior
\begin{equation}
    q(\Z) = \N(\langle\Z\rangle, \Sigma_{\Z}^{-1})
\end{equation}
where
\begin{align}
    \Sigma_{\mathbf{Z}}^{-1} & = \tauE \Ik + \langle\psi\rangle\langle\mathbf{V}\mathbf{V}^\top\rangle
        \\ \langle\mathbf{Z}\rangle & = \left(\langle\tau\rangle\sum_m^M \mathbf{H}(m,s^{\rm (m)}) + \langle\psi\rangle\mathbf{Y}\langle\mathbf{V}\rangle\right)\Sigma_{\mathbf{Z}}
\end{align}

\subsection{Mean Field Approximation of $\mathbf{A}$}
\label{sec:inf_A}

As stated in the mean-field inference procedure, we begin rewriting Eq. (\ref{eq:lnqopt}) as:
\begin{equation}
\label{eq:E_a}
    \ln q(\mathbf{A}^\M) = \E\left[\ln p(\mathbf{Z}|\mathbf{W}^{\M}\mathbbm{1}(s^\M = 0),\mathbf{A}^{\M}\mathbbm{1}(s^\M = 1),\tau) + \ln p(\mathbf{\tilde{X}}^{\M\top}\mathbf{A}^{\M}|\boldsymbol{\gamma}^\M, \boldsymbol{\delta}^\M)\right].
\end{equation}

Thus, we can develop the first term as presented in Eq. (\ref{eq:zy}) as:
\begin{equation}
\begin{split}
\label{eq:z_para_a}
    & \ln p(\mathbf{Z}|\mathbf{W}^{\M}\mathbbm{1}(s^\M = 0),\mathbf{A}^{\M}\mathbbm{1}(s^\M = 1),\tau) \\
    & = \sumn\ln\N\left(\summ\left(\left(\mathbf{x}_{n,:}\Wm^\top\right)\mathbbm{1}(s^m = 0) + \left(\mathbf{x}_{n,:}\Xsmt\mathbf{A}^{\rm (m)}\right)\mathbbm{1}(s^m = 1)\right), \tau\right) \\
    & = \sumn \Bigg(-\unmed\lnpi + \unmed\ln \tau - \frac{\tau}{2}\left(\Zn- \summ \mathbf{P}(m,s^{\rm (m)})_{n,:}\right) \left(\Zn - \summ \mathbf{P}(m,s^{\rm (m)})_{n,:}\right)^\top\Bigg).
\end{split}
\end{equation}

As we will further apply the expectation over $\mathbf{A}$, we can rewrite Eq. (\ref{eq:z_para_a}) as:
\begin{equation}
\begin{split}
\label{eq:final_aa}
    & \ln p(\mathbf{Z}|\mathbf{W}^{\M}\mathbbm{1}(s^\M = 0),\mathbf{A}^{\M}\mathbbm{1}(s^\M = 1),\tau) \\
    & = \sumn\Bigg(-\frac{\tau}{2}\Bigg(-2\Zn\summ \mathbf{P}(m,s^{\rm (m)})_{n,:}^\top + \summ \mathbf{P}(m,s^{\rm (m)})_{n,:}\summ \mathbf{P}(m,s^{\rm (m)})_{n,:}^\top\Bigg)\Bigg) \\
    & = \sumn\Bigg(-\frac{\tau}{2}\Bigg(-2\Zn\summ \mathbf{P}(m,s^{\rm (m)})_{n,:}^\top + \summ\sum_{m'=1}^M \mathbf{P}(m,s^{\rm (m)})_{n,:}\mathbf{P}(m',s^{\rm (m')})_{n,:}^\top\Bigg)\Bigg) \\
    & = \sumn\Bigg(-\frac{\tau}{2}\Bigg(-2\Zn\summ \mathbf{P}(m,s^{\rm (m)})_{n,:}^\top \\ & + \summ\sum_{m'=1}^M \Bigg(\Bigg(\Xnm\Wm\mathbf{A}^{(m')\top}\mathbf{\tilde{X}}^{(m')}\mathbf{x}_{\rm n,:}^{(m')\top}\Bigg)\mathbbm{1}(s^{(m)} = 0\vert s^{(m')} = 1)\Bigg)\\
    & + \Bigg(\Xnm\Wm\mathbf{W}^{(m')\top}\mathbf{x}_{\rm n,:}^{(m')\top}\Bigg)\mathbbm{1}(s^{(m)} = 0\vert s^{(m')} = 0)\Bigg) \\ & + \Bigg(\Xnm\Xsmt\mathbf{A}^{(m)}\mathbf{A}^{(m')\top}\mathbf{\tilde{X}}^{(m')}\mathbf{x}_{\rm n,:}^{(m')\top}\Bigg)\mathbbm{1}(s^{(m)} = 1\vert s^{(m')} = 1)\\
    & + \Bigg(\Xnm\Xsmt\mathbf{A}^{(m)}\mathbf{W}^{(m')}\mathbf{x}_{\rm n,:}^{(m')\top}\Bigg)\mathbbm{1}(s^{(m)} = 1\vert s^{(m')} = 0)\Bigg)\Bigg) + \const.
\end{split}
\end{equation}

Also, if we disassemble the inference of each $\mathbf{A}^{(m)}$, we can rewrite Eq. (\ref{eq:final_aa}) as:
\begin{equation}
\begin{split}
    & \ln p(\mathbf{Z}|\mathbf{W}^{\M}\mathbbm{1}(s^\M = 0),\mathbf{A}^{\M}\mathbbm{1}(s^\M = 1),\tau) \\
    & = \sumn\Bigg(-\frac{\tau}{2}\Bigg(-2\Zn\summ \mathbf{A}^{(m)\top}\Xsm\mathbf{x}_{\rm n,:}^{(m)\top} \\ 
    & + \summ\sum_{m'=1}^M \Bigg(\Bigg(\Xnm\Xsmt\mathbf{A}^{(m)}\mathbf{A}^{(m')\top}\mathbf{\tilde{X}}^{(m')}\mathbf{x}_{\rm n,:}^{(m')\top}\Bigg)\mathbbm{1}(s^{(m)} = 1\vert s^{(m')} = 1) \\ 
    & + \Bigg(\Xnm\Xsmt\mathbf{A}^{(m)}\mathbf{W}^{(m')}\mathbf{x}_{\rm n,:}^{(m')\top}\Bigg)\mathbbm{1}(s^{(m)} = 1\vert s^{(m')} = 0)\Bigg)\Bigg)\Bigg) + \const.
\end{split}
\end{equation}

Also, as it is not possible to calculate the probability over a matrix, we impose a summation over $K$ to define the final probability of each $\mathbf{a}_{\rm :,k}^{(m)}$ as:
\begin{equation}
\begin{split}
\label{eq:last_z}
    & \ln p(\mathbf{Z}|\mathbf{W}^{\M}\mathbbm{1}(s^\M = 0),\mathbf{A}^{\M}\mathbbm{1}(s^\M = 1),\tau) \\
    & = \sumn\sumk\Bigg(-\frac{\tau}{2}\Bigg(-2 z_{\rm n,k}\summ \mathbf{x}_{\rm n,:}^{(m)}\Xsmt\mathbf{a}_{\rm :,k}^{(m)} \\
    & + \summ\sum_{m'=1}^M \Bigg( \Bigg(\mathbf{a}_{\rm :,k}^{(m)\top}\Xsm\mathbf{x}_{\rm n,:}^{(m)\top}\mathbf{x}_{\rm n,:}^{(m')}\mathbf{\tilde{X}}^{(m')\top}\mathbf{a}_{\rm :,k}^{(m')}\Bigg)\mathbbm{1}(s^{(m')} = 1)\\
    & + \Bigg(\mathbf{a}_{\rm :,k}^{(m)\top}\Xsm\mathbf{x}_{\rm n,:}^{(m)\top}\mathbf{x}_{\rm n,:}^{(m')}\mathbf{w}_{\rm k,:}^{(m')\top}\Bigg)\mathbbm{1}(s^{(m')} = 0)\Bigg)\Bigg) + \const.
\end{split}
\end{equation}

Also, as in Eq. (\ref{eq:last_z}), we will develop the second term of Eq. (\ref{eq:E_a}) for each $\mathbf{A}^{(m)}$. Thus, we can rewrite Eq. (\ref{eq:E_a}) as:
\begin{equation}
\begin{split}
\label{eq:term2_a}
    & \ln p(\mathbf{\tilde{X}}^{(m)\top}\mathbf{A}^{(m)}\vert\boldsymbol{\gamma}^{(m)},\boldsymbol{\delta}^{(m)}) = \sumk \ln \N\left(\mathbf{\tilde{X}}^{(m)\top}\mathbf{A}^{(m)}\vert 0, \delta_k^{(m)-1}\Lambda_{\boldsymbol{\gamma}^{(m)}}^{-1}\right) \\
    & = \sumk \left(-\unmed\lnpi + \unmed\ln(\delta_k^{(m)}\Lambda_{\boldsymbol{\gamma}^{(m)}}) - \frac{\delta_k^{(m)}}{2}\left(\mathbf{a}^{(m)\top}_{\rm :,k}\mathbf{\tilde{X}}^{(m)}\Lambda_{\boldsymbol{\gamma^{(m)}}}\mathbf{\tilde{X}}^{(m)\top}\mathbf{a}^{(m)}_{\rm :,k}\right)\right) \\
    & = \sumk\sumd\left(- \frac{(\delta_k^{(m)}\gamma_d^{(m)})}{2}\mathbf{a}^{(m)\top}_{\rm :,k}\mathbf{\tilde{x}}_{\rm :,d}^{(m)}\mathbf{\tilde{x}}_{\rm :,d}^{(m)\top}\mathbf{a}^{(m)}_{\rm :,k}\right)+ \const.
\end{split}
\end{equation}

Now, if we include both Eq. (\ref{eq:last_z}) and (\ref{eq:term2_a}) within Eq. (\ref{eq:E_a}) and group the variables, we obtain:
\begin{equation}
\begin{split}
\label{eq:final_aa}
    &\ln q(\mathbf{a}_{:,k}^{(m)}) = \E\Bigg[\sumk\Bigg(\mathbf{a}_{\rm :,k}^{(m)\top}\left(-\frac{\tau}{2}\Xsm\mathbf{x}_{\rm n,:}^{(m)\top}\mathbf{x}_{\rm n,:}^{(m)}\mathbf{\tilde{X}}^{(m)\top} - \frac{(\delta_k^{(m)}\gamma_d^{(m)})}{2}\mathbf{\tilde{x}}_{\rm :,d}^{(m)}\mathbf{\tilde{x}}_{\rm :,d}^{(m)\top}\right)\mathbf{a}_{\rm :,k}^{(m)} \\
    & + \Bigg(\tau\mathbf{\tilde{X}}^{(m)}\XmT\mathbf{z}_{\rm :,k} - \tau\mathbf{\tilde{X}}^{(m)}\XmT\sum_{m' \neq m}^M\left(\mathbf{x}_{\rm n,:}^{(m')}\mathbf{W}^{(m')\top}\right)\mathbbm{1}(s^{(m')} = 0)\\
    &- \tau\mathbf{\tilde{X}}^{(m)}\XmT\sum_{m' \neq m}^M\left(\mathbf{x}_{\rm n,:}^{(m')}\mathbf{\tilde{X}}^{(m')\top}\mathbf{a}_{\rm :,k}^{(m')}\right)\mathbbm{1}(s^{(m')} = 1)\Bigg)\mathbf{a}_{\rm :,k}^{(m)}\Bigg)\Bigg].
\end{split}
\end{equation}

Finally, if we apply the expectation over Eq. (\ref{eq:final_aa}) and identify terms, we can parameterize the approximated posterior $q(\mathbf{a}_{\rm :,k}^{(m)})$ as:
\begin{equation}
    q(\mathbf{a}_{\rm :,k}^{(m)}) = \N(\langle\mathbf{a}_{\rm :,k}^{(m)}\rangle, \Sigma_{\mathbf{a}_{\rm :,k}^{(m)}}^{-1})
\end{equation}
where 
\begin{align}
    \Sigma_{a_{:,\rm k}}^{(m)-1} & = \tauE \Xsm\XmT\Xm\Xsmt + \langle\akm \rangle\Xsm \Lambda_{\langle\gamm\rangle}\Xsmt
        \\ \langle\mathbf{a}_{:,\rm k}^{\rm (m)}\rangle & = \tauE\Sigma_{a_{:,\rm k}}^{\rm (m)}\Xsm\XmT\left[\langle\mathbf{z}_{:,\rm k}\rangle - \sum_{m' \neq m}^{M}\mathbf{H}(m',s^{\rm (m')})_{:,k}\right],
\end{align}
where $\mathbf{H}(m',s^{\rm (m')})_{:,k}$ is defined at Eq. (\ref{eq:Hh}).

\subsection{Mean Field Approximation of $\mathbf{W}$}

The development of $\mathbf{W}$ is analog to the one presented for $\mathbf{A}$ in the previous Section \ref{sec:inf_A}. Thus, if we rewrite Eq. (\ref{eq:lnqopt}) for $\mathbf{W}$, we obtain:
\begin{equation}
    \ln q(\mathbf{W}^\M) = \E\left[\ln p(\mathbf{Z}|\mathbf{W}^{\M}\mathbbm{1}(s^\M = 0),\mathbf{A}^{\M}\mathbbm{1}(s^\M = 1),\tau) + \ln p(\mathbf{W}^{\M}|\boldsymbol{\gamma}^\M, \boldsymbol{\delta}^\M)\right].
\end{equation}

Thus, the arithmetical development of both terms will lead to similar expressions as depicted in Eq. (\ref{eq:last_z}) and (\ref{eq:term2_a}). Hence, the final approximated posterior of $\mathbf{w}_{\rm k,:}^{(m)}$ will look as:
\begin{equation}
    q(\mathbf{w}_{\rm k,:}^{(m)}) = \N(\langle\mathbf{w}_{\rm k,:}^{(m)}\rangle, \Sigma_{\mathbf{w}_{\rm k,:}^{(m)}}^{-1})
\end{equation}
where
\begin{align}
    \Sigma_{\Wk}^{(m)-1} &= \tauE \XmT\Xm + \langle\akm \rangle \Lambda_{\langle\gamm\rangle}
        \\ \langle \Wkm\rangle &= \tauE \left[\langle\mathbf{z}_{:,\rm k}\rangle - \sum_{m' \neq m}^{M}\mathbf{H}(m',s^{\rm (m')})^\top\right]\Xm\Sigma_{\Wk}^{\rm (m)}
\end{align}

\subsection{Mean Field Approximation of $\tau$}

As the mean-field approximation states, we adapt Eq. (\ref{eq:lnqopt}) for $\tau$ as:
\begin{equation}
\label{eq:E_tau}
    \ln q(\tau) = \E\left[\ln p(\mathbf{Z}|\mathbf{W}^{\M}\mathbbm{1}(s^\M = 0),\mathbf{A}^{\M}\mathbbm{1}(s^\M = 1),\tau) + \ln p(\tau)\right].
\end{equation}

The first term can be developed similarly as Eq. (\ref{eq:final_aa}); that is
\begin{equation}
\begin{split}
    & \ln p(\mathbf{Z}|\mathbf{W}^{\M}\mathbbm{1}(s^\M = 0),\mathbf{A}^{\M}\mathbbm{1}(s^\M = 1),\tau) \\
    & = \sumn\Bigg(\unmed\ln\tau -\frac{\tau}{2}\Bigg(\Zn\ZnT -2\Zn\summ \mathbf{P}(m,s^{\rm (m)})_{n,:}^\top + \summ \mathbf{P}(m,s^{\rm (m)})_{n,:}\summ \mathbf{P}(m,s^{\rm (m)})_{n,:}^\top\Bigg)\Bigg) \\
    & = \sumn\Bigg(\unmed\ln\tau -\frac{\tau}{2}\Bigg(\Zn\ZnT -2\Zn\summ \mathbf{P}(m,s^{\rm (m)})_{n,:}^\top + \summ\sum_{m'=1}^M \mathbf{P}(m,s^{\rm (m)})_{n,:}\mathbf{P}(m',s^{\rm (m')})_{n,:}^\top\Bigg)\Bigg) \\
    & = \sumn\Bigg(\unmed\ln\tau -\frac{\tau}{2}\Bigg(\Zn\ZnT-2\Zn\summ \mathbf{P}(m,s^{\rm (m)})_{n,:}^\top \\
    & + \summ\sum_{m'=1}^M \Bigg(\Bigg(\Xnm\Wm\mathbf{A}^{(m')\top}\mathbf{\tilde{X}}^{(m')}\mathbf{x}_{\rm n,:}^{(m')\top}\Bigg)\mathbbm{1}(s^{(m)} = 0\vert s^{(m')} = 1)\Bigg)\\
    & + \Bigg(\Xnm\Wm\mathbf{W}^{(m')\top}\mathbf{x}_{\rm n,:}^{(m')\top}\Bigg)\mathbbm{1}(s^{(m)} = 0\vert s^{(m')} = 0)\Bigg) \\
    & + \Bigg(\Xnm\Xsmt\mathbf{A}^{(m)}\mathbf{A}^{(m')\top}\mathbf{\tilde{X}}^{(m')}\mathbf{x}_{\rm n,:}^{(m')\top}\Bigg)\mathbbm{1}(s^{(m)} = 1\vert s^{(m')} = 1)\\
    & + \Bigg(\Xnm\Xsmt\mathbf{A}^{(m)}\mathbf{W}^{(m')}\mathbf{x}_{\rm n,:}^{(m')\top}\Bigg)\mathbbm{1}(s^{(m)} = 1\vert s^{(m')} = 0)\Bigg)\Bigg) + \const.
\end{split}
\end{equation}

To apply further expectations over the model variables, we will slightly operate over the whole equation. Thus, the final equation will look as follows
\begin{equation}
\label{eq:term1_tau}
\begin{split}
    & \ln p(\mathbf{Z}|\mathbf{W}^{\M}\mathbbm{1}(s^\M = 0),\mathbf{A}^{\M}\mathbbm{1}(s^\M = 1),\tau) \\
    & = \sumn\sumk\Bigg(\unmed\ln\tau -\frac{\tau}{2}\Bigg(z_{\rm n,k}^{2}-2 z_{\rm n,k}\summ \mathbf{x}_{\rm n,:}^{(m)}\Xsmt\mathbf{a}_{\rm :,k}^{(m)} \\
    & + \summ\sum_{m'=1}^M \Bigg( \Bigg(\mathbf{a}_{\rm :,k}^{(m)\top}\Xsm\mathbf{x}_{\rm n,:}^{(m)\top}\mathbf{x}_{\rm n,:}^{(m')}\mathbf{\tilde{X}}^{(m')\top}\mathbf{a}_{\rm :,k}^{(m')}\Bigg)\mathbbm{1}(s^{(m)} = 1\vert s^{(m')} = 1)\\
    & + \Bigg(\mathbf{a}_{\rm :,k}^{(m)\top}\Xsm\mathbf{x}_{\rm n,:}^{(m)\top}\mathbf{x}_{\rm n,:}^{(m')}\mathbf{w}_{\rm k,:}^{(m')\top}\Bigg)\mathbbm{1}(s^{(m)} = 1\vert s^{(m')} = 0)\\
    & + \Bigg(\mathbf{w}_{\rm k,:}^{(m)}\mathbf{x}_{\rm n,:}^{(m)\top}\mathbf{x}_{\rm n,:}^{(m')}\mathbf{w}_{\rm k,:}^{(m')\top}\Bigg)\mathbbm{1}(s^{(m)} = 0\vert s^{(m')} = 0)\\
    & + \Bigg(\mathbf{w}_{\rm k,:}^{(m)}\mathbf{x}_{\rm n,:}^{(m)\top}\mathbf{x}_{\rm n,:}^{(m')}\mathbf{\tilde{X}}^{(m')\top}\mathbf{a}_{\rm :,k}^{(m')}\Bigg)\mathbbm{1}(s^{(m)} = 0\vert s^{(m')} = 1) \Bigg)\Bigg) + \const. \\
    & = \sumn\sumk\Bigg(\unmed\ln\tau -\frac{\tau}{2}\Bigg(z_{\rm n,k}^{2}-2 z_{\rm n,k}\summ \mathbf{x}_{\rm n,:}^{(m)}\Xsmt\mathbf{a}_{\rm :,k}^{(m)} \\
    & + \summ\sum_{m' \neq m}^M \Bigg( \Bigg(\mathbf{a}_{\rm :,k}^{(m)\top}\Xsm\mathbf{x}_{\rm n,:}^{(m)\top}\mathbf{x}_{\rm n,:}^{(m')}\mathbf{\tilde{X}}^{(m')\top}\mathbf{a}_{\rm :,k}^{(m')}\Bigg)\mathbbm{1}(s^{(m)} = 1\vert s^{(m')} = 1)\\
    & + \Bigg(\mathbf{w}_{\rm k,:}^{(m)}\mathbf{x}_{\rm n,:}^{(m)\top}\mathbf{x}_{\rm n,:}^{(m')}\mathbf{w}_{\rm k,:}^{(m')\top}\Bigg)\mathbbm{1}(s^{(m)} = 0\vert s^{(m')} = 0) \Bigg)\\
    & + \summ\sum_{m' = 1}^M \Bigg( \Bigg(\mathbf{a}_{\rm :,k}^{(m)\top}\Xsm\mathbf{x}_{\rm n,:}^{(m)\top}\mathbf{x}_{\rm n,:}^{(m')}\mathbf{w}_{\rm k,:}^{(m')\top}\Bigg)\mathbbm{1}(s^{(m)} = 1\vert s^{(m')} = 0)\\
    & + \Bigg(\mathbf{w}_{\rm k,:}^{(m)}\mathbf{x}_{\rm n,:}^{(m)\top}\mathbf{x}_{\rm n,:}^{(m')}\mathbf{\tilde{X}}^{(m')\top}\mathbf{a}_{\rm :,k}^{(m')}\Bigg)\mathbbm{1}(s^{(m)} = 0\vert s^{(m')} = 1) \Bigg) \\
    &+ \summ\sum_{m' = m}^M \Bigg( Tr\Bigg(\Xsm\mathbf{x}_{\rm n,:}^{(m)\top}\mathbf{x}_{\rm n,:}^{(m')}\mathbf{\tilde{X}}^{(m')\top}\mathbf{a}_{\rm :,k}^{(m')}\mathbf{a}_{\rm :,k}^{(m)\top}\Bigg)\mathbbm{1}(s^{(m)} = 1)\\
    & + \Bigg(\mathbf{x}_{\rm n,:}^{(m)\top}\mathbf{x}_{\rm n,:}^{(m')}\mathbf{w}_{\rm k,:}^{(m')\top}\mathbf{w}_{\rm k,:}^{(m)}\Bigg)\mathbbm{1}(s^{(m)} = 0)\Bigg)\Bigg) + \const. 
\end{split}
\end{equation}

Note that the Trace operator was applied over two final terms to apply the expectation over them. That is, having the variables under expectation together as: $\langle\mathbf{w}_{\rm k,:}^{(m)\top}\mathbf{w}_{\rm k,:}^{(m)}\rangle$ and $\langle\mathbf{a}_{\rm :,k}^{(m)}\mathbf{a}_{\rm :,k}^{(m)\top}\rangle$.

Now, the second term of Eq. (\ref{eq:E_tau}) can be easily developed following the general formula of the gamma distribution:
\begin{equation}
\begin{split}
\label{eq:term2_tau}
    \ln p(\tau) = \ln \Gamma(\alpha_0^\tau,\beta_0^\tau) = -\bcerotau\taus + (\acerotau -1)\lntau + \const.
\end{split}
\end{equation}

Now, if we include both Eq. (\ref{eq:term1_tau}) and (\ref{eq:term2_tau}) within Eq. (\ref{eq:E_tau}), apply the expectation and identify terms, we can parameterize the approximated posterior distribution of $\tau$ as
\begin{equation}
    q(\tau) = \Gamma(\alpha^\tau, \beta^\tau),
\end{equation}
where
\begin{align}
    \alpha^{\tau} &= \frac{NK}{2} + \alpha_0^{\tau}\\
    \beta^{\tau} &= \beta_0^{\tau} + \unmed\sum_n^N\langle\mathbf{Z}\mathbf{Z}^\top\rangle_{n,n} - Tr\left(\langle\mathbf{Z}\rangle\sum_m^{M}\mathbf{H}(m,s^{\rm (m)})^\top\right) \\ & + \unmed Tr\left(\sum_m^M\sum_{m'}^M \mathbf{H}(m,s^{\rm (m)})\mathbf{H}(m',s^{(m')})^\top\right)
\end{align}

\subsection{Mean Field Approximation of $\delta$}
\label{sec:inf_delta}

We redefine Eq. (\ref{eq:lnqopt}) with $\boldsymbol{\delta}$ as:
\begin{equation}
\label{eq:E_delta}
    \ln q(\boldsymbol{\delta}^\M) = \E\left[\ln \left(p(\mathbf{W}^{\M}|\boldsymbol{\gamma}^\M, \boldsymbol{\delta}^\M)^{\mathbbm{1}(s^\M = 0)} p(\mathbf{\tilde{X}}^{\M\top}\mathbf{A}^{\M}|\boldsymbol{\gamma}^\M, \boldsymbol{\delta}^\M)^{\mathbbm{1}(s^\M = 1)}\right) + \ln p(\boldsymbol{\delta}^\M)\right].
\end{equation}

Note that, the final distribution of each $\boldsymbol{\delta}^{(m)}$ will depend on if the view is fat or not. That is, as presented in previous Eq. (\ref{eq:E_delta}), we will use either $p(\mathbf{W}^{\M}|\boldsymbol{\gamma}^\M, \boldsymbol{\delta}^\M)$ or $p(\mathbf{\tilde{X}}^{\M\top}\mathbf{A}^{\M}|\boldsymbol{\gamma}^\M, \boldsymbol{\delta}^\M)$ if the view will work over the primal or dual space, respectively. However, we will provide a compact and general solution that enables both cases.

First, we will develop the first term for the primal and dual case. For the primal case, i.e., $\mathbbm{1}(s^\M = 0)$, first term can be developed as:
\begin{equation}
\begin{split}
\label{eq:term1_primal_delta}
    & \ln p(\mathbf{W}^{(m)}\vert\boldsymbol{\gamma}^{(m)},\boldsymbol{\delta}^{(m)}) = \sumk \ln \N\left(\mathbf{W}^{(m)}\vert 0, \delta_k^{(m)-1}\Lambda_{\boldsymbol{\gamma}^{(m)}}^{-1}\right) \\
    & = \sumk \left(-\unmed\lnpi + \unmed\ln(\delta_k^{(m)}\Lambda_{\boldsymbol{\gamma}^{(m)}}) - \frac{\delta_k^{(m)}}{2}\left(\mathbf{w}_{\rm k,:}\Lambda_{\boldsymbol{\gamma^{(m)}}}\mathbf{w}_{\rm k,:}^\top\right)\right) \\
    & = \sumk\sumd\left(\unmed\ln(\delta_k^{(m)}\gamma_d^{(m)})- \frac{(\delta_k^{(m)}\gamma_d^{(m)})}{2} w_{\rm k,d}^{(m)2}\right)+ \const.
\end{split}
\end{equation}

Also, if we consider the dual case, i.e., $\mathbbm{1}(s^\M = 1)$, we will develop the first term as:
\begin{equation}
\begin{split}
\label{eq:term1_dual_delta}
    & \ln p(\mathbf{\tilde{X}}^{(m)\top}\mathbf{A}^{(m)}\vert\boldsymbol{\gamma}^{(m)},\boldsymbol{\delta}^{(m)}) = \sumk \ln \N\left(\mathbf{\tilde{X}}^{(m)\top}\mathbf{A}^{(m)}\vert 0, \delta_k^{(m)-1}\Lambda_{\boldsymbol{\gamma}^{(m)}}^{-1}\right) \\
    & = \sumk \left(-\unmed\lnpi + \unmed\ln(\delta_k^{(m)}\Lambda_{\boldsymbol{\gamma}^{(m)}}) \frac{\delta_k^{(m)}}{2}\left(\mathbf{a}^{(m)\top}_{\rm :,k}\mathbf{\tilde{X}}^{(m)}\Lambda_{\boldsymbol{\gamma^{(m)}}}\mathbf{\tilde{X}}^{(m)\top}\mathbf{a}^{(m)}_{\rm :,k}\right)\right) \\
    & = \sumk\sumd\left(\unmed\ln(\delta_k^{(m)}\gamma_d^{(m)})- \frac{(\delta_k^{(m)}\gamma_d^{(m)})}{2} \mathbf{a}^{(m)\top}_{\rm :,k}\mathbf{\tilde{x}}_{\rm :,d}^{(m)}\mathbf{\tilde{x}}_{\rm :,d}^{(m)\top}\mathbf{a}^{(m)}_{\rm :,k}  \right)+ \const.
\end{split}
\end{equation}

Also, if we develop the second term for each $\delta^{(m)}$, we will obtain:
\begin{equation}
    \ln p(\boldsymbol{\delta}^{(m)}) = \Gamma(\boldsymbol{\alpha}_0^{(m)\delta},\boldsymbol{\beta}_{0}^{(m)\delta}) = \sumk\left(-\beta_{0 k}^{(m)\delta}\delta_k^{(m)} + (\alpha_{0 k}^{(m)\delta} -1)\ln \delta_k^{(m)} \right)
\end{equation}

Now, if include both terms in Eq. (\ref{eq:E_delta}) and apply expectations, we will obtain de final posterior approximation of each $k$ element of $\mathbf{\delta}^{(m)}$ as:
\begin{equation}
    q(\delta_k^{(m)}) = \Gamma(\alpha_{\rm k}^{\delta(m)},\beta_{\rm k}^{\delta(m)}),
\end{equation}
where
\begin{equation}
    \boldsymbol{\alpha}^{\delta(m)} = \frac{D^{\rm (m)}}{2} + \alpha_0^{\delta(m)},
\end{equation}
and
\begin{equation}
\begin{split}
    \beta_{\rm k}^{\delta(m)} =& \beta_0^{\delta(m)} + \unmed\sum_{\rm d}^{D^{\rm (m)}}\Bigg[\langle\gamma_d^{\rm (m)}\rangle\Bigg(\langle w_{\rm k,d}^{(m)^{2}}\rangle\mathbbm{1}(s^{\rm (m)} = 0) \\
    &+ \left(\mathbf{\tilde{X}}_{:,\rm d}^{(m)\top}\langle\mathbf{a}_{:,\rm k}^{\rm (m)}\mathbf{a}_{:,\rm k}^{(m)\top}\rangle\tilde{X}_{:,\rm d}^{\rm (m)}\right)\mathbbm{1}(s^{\rm (m)} = 1)\Bigg)\Bigg]
\end{split}
\end{equation}

\subsection{Mean Field Approximation of $\gamma$}

The inference of each $\gamma_d^{(m)}$ will follow an analog development as presented in the previous Section \ref{sec:inf_delta}. That is, the final approximated posterior will look as follows:
\begin{equation}
    q(\gamma_d^{(m)}) = \Gamma(\alpha_{\rm d}^{\gamma(m)},\beta_{\rm d}^{\gamma(m)}),
\end{equation}
where
\begin{equation}
    \boldsymbol{\alpha}^{\gamma(m)} = \frac{K}{2} + \alpha_0^{\gamma(m)},
\end{equation}
and
\begin{equation}
\begin{split}
    \beta_{d\rm }^{\gamma(m)} &= \beta_0^{\gamma(m)} + \unmed\sum_{\rm k}^{K}\Bigg[\langle\delta_{\rm k}^{\rm (m)}\rangle\Bigg(\langle w_{\rm k,d}^{(m)^{2}}\rangle\mathbbm{1}(s^{\rm (m)} = 0) \\ &+ \left(\mathbf{\tilde{X}}_{:,\rm d}^{(m)\top}\langle\mathbf{a}_{:,\rm k}^{\rm (m)}\mathbf{a}_{:,\rm k}^{(m)\top}\rangle\tilde{X}_{:,\rm d}^{\rm (m)}\right)\mathbbm{1}(s^{\rm (m)} = 1)\Bigg)\Bigg].
\end{split}
\end{equation}

\subsection{Mean Field Approximation of $\mathbf{V}$}

For $\mathbf{V}$ case, Eq. (\ref{eq:lnqopt}) can be rewrote as:
\begin{equation}
\label{eq:E_v}
    \ln q(\mathbf{V}) = \E\left[\ln p(\mathbf{Y}|\mathbf{Z}, \mathbf{V},\psi) + \ln p(\mathbf{V}|\boldsymbol{\omega}) \right].
\end{equation}

In a similar fashion as Eq. (\ref{eq:yz}), we can develop first term as:
\begin{equation}
\label{eq:term1_v}
\begin{split}
    \ln p(\mathbf{Y}|\mathbf{Z}, \mathbf{V},\psi) & = \sum_{n=1}^N \ln \N(\mathbf{z}_{n,:}\mathbf{V}^\top,\psi) \\
    & = \sum_{n=1}^{N}\left(-\unmed\lnpi + \unmed\ln \psi -\frac{\psi}{2}(\yn - \mathbf{z}_{n,:}\mathbf{V}^\top)(\yn - \mathbf{z}_{n,:}\mathbf{V}^\top)^\top\right) \\
    & = \sum_{n=1}^{N}\left(-\frac{\psi}{2}\left(-2\Zn\VT\YnT + + \Zn\VT\V\ZnT\right)\right) + \const. \\
    & =\sumn\sumc \left(-\frac{\psi}{2}\left(-2\mathbf{v}_{\rm c,:}\mathbf{z}_{\rm n,:}^\top y_{\rm n,c} + \mathbf{v}_{\rm c,:}\mathbf{z}_{\rm n,:}^\top\mathbf{z}_{\rm n,:}\mathbf{v}_{\rm c,:}^\top\right)\right) + \const.
\end{split}
\end{equation}

Also, the second term will follow a similar pattern as Eq. (\ref{eq:term2_a}). That is, we can rewrite the second term as:
\begin{equation}
\label{eq:term2_v}
    \ln p(\mathbf{V}|\boldsymbol{\omega}) = \sumc\left(-\unmed\mathbf{v}_{\rm c,:}\Lambda_{\boldsymbol{\omega}}\mathbf{v}_{\rm c,:}^{\top}\right) + \const.
\end{equation}

Now, if we include both Eq. (\ref{eq:term1_v}) and (\ref{eq:term2_v}) in Eq. (\ref{eq:E_v}) and apply expectations, we will obtain the final form of the approximated posterior of $\mathbf{v}$ as:
\begin{equation}
    \V = \N \p*{\langle\V\rangle, \Sigma_{\V}},
\end{equation}
where
\begin{align}
    \Sigma_{\V}^{-1} &= \Lambda_{\ang{\bet}} + \ang{\psi}\ang{\ZT\Z} \\ \langle\V\rangle &= \ang{\psi} \YT \ang{\Z} \Sigma_{\V}
\end{align}

\subsection{Mean Field Approximation of $\psi$}

For the inference of the output noise, we rewrite Eq. (\ref{eq:lnqopt}) as:
\begin{equation}
\label{eq:E_psi}
    \ln q(\psi) = \E\left[\ln p(\mathbf{Y}|\mathbf{Z}, \mathbf{V},\psi) + \ln p(\psi)\right].
\end{equation}

The first term can be developed as:
\begin{equation}
\label{eq:term1_psi}
\begin{split}
    &\ln p(\mathbf{Y}|\mathbf{Z}, \mathbf{V},\psi) = \sumn\ln\N\left(\mathbf{z}_{\rm n,:}\mathbf{V}, \psi^{-1}\Ic\right) \\
    & = \sumn\left(\unmed\ln\psi - \unmed\psi\left(\Yn\YnT - 2\Zn\VT\YnT + Tr\left(\ZnT\Zn\VT\V\right)\right)\right) + \const.
\end{split}
\end{equation}

Then, following an analog idea as Eq. (\ref{eq:term2_tau}), we can develop second term as:
\begin{equation}
\begin{split}
\label{eq:term2_psi}
    \ln p(\psi) = \ln \Gamma(\alpha_0^\psi,\beta_0^\psi) = -\beta_0^\psi\psi + (\alpha_0^\psi -1)\ln \psi + \const.
\end{split}
\end{equation}

Thus, if we integrate both Eq. (\ref{eq:term1_psi}) and (\ref{eq:term2_psi}) within Eq. (\ref{eq:E_psi}) and apply expectations, we will obtain:
\begin{equation}
    q(\psi) = \Gamma(\alpha^\psi, \beta^\psi),
\end{equation}
where
\begin{align}
    \alpha^{\psi} & = \frac{N C}{2} + \alpha^{\psi}_0 \\ \beta^{\psi} & = \beta^{\psi}_0 + \frac{1}{2}\sum_n^N\mathbf{y}_{\rm n,:}\YnT - Tr\llav{\Y  \ang{\V} \ang{\ZT}} + \frac{1}{2} Tr\llav{\ang{\VT\V} \ang{\ZT\Z}}
\end{align}

\subsection{Mean Field Approximation of $\omega$}

In the case of $\boldsymbol{\omega}$, we can rewrite Eq. (\ref{eq:lnqopt}) as:
\begin{equation}
\label{eq:E_omega}
    \ln q(\boldsymbol{\omega}) = \E\left[\ln p(\mathbf{V}|\boldsymbol{\omega}) + \ln p(\boldsymbol{\omega})\right].
\end{equation}

Following a similar pathway as in Eq. (\ref{eq:term2_v}), first term can be expressed as:
\begin{equation}
\label{eq:term1_omega}
    \ln p(\mathbf{V}|\boldsymbol{\omega}) = \sumk\sumc\left(\unmed\ln \omega_k - \unmed\omega_k v_{\rm c,k}^2\right),
\end{equation}
while the second term as:
\begin{equation}
\label{eq:term2_omega}
    \ln p(\boldsymbol{\omega}) = \sumk\left(-\beta_{0 k}^\omega \omega_k + (\alpha_{0 k}-1)\ln \omega_k\right).
\end{equation}

Hence, if we include both terms into Eq. (\ref{eq:E_omega}) and apply expectations, we obtain:
\begin{equation}
    q(\omega_k) = \Gamma(\alpha_k^\omega, \beta_k^\omega),
\end{equation}
where
\begin{align}
    \alpha_{\rm k}^{\omega} &= \frac{C}{2} + \alpha^{\omega}_0 \\ 
    \beta_{\rm k}^{\omega} &= \beta^{\omega}_0 + \frac{1}{2}\ang{\VkT\Vk}
\end{align}

\subsection{Mean Field Approximation of $\mathbf{Y}$}

Finally, we will parameterize the posterior distribution of the soft output view, i.e., $q(\mathbf{Y})$. To do so, we redefine Eq. (\ref{eq:lnqopt}) as:
\begin{equation}
\label{eq:E_y_ant}
    \ln q(\mathbf{Y}) = \E\left[\ln p(\mathbf{T}|\mathbf{Y}) + \ln p(\mathbf{Y}|\mathbf{Z}, \mathbf{V},\psi)\right].
\end{equation}

Furthermore, as explained in Section \ref{sec:Inference_BALDUR}, we lower bound the first-order Taylor serie of $p(\mathbf{T}|\mathbf{Y})$ to achieve a final distribution capable of conjugate with the Gaussian distribution that follows $p(\mathbf{Y}|\mathbf{Z}, \mathbf{V},\psi)$. That is, we obtain:
\begin{equation}
    \ln p(\mathbf{t}_{\rm n,:} = 1|\mathbf{y}_{\rm n,:}) = 
    e^{\mathbf{y}_{\rm n,:}}\sigma(-\mathbf{y}_{\rm n,:}) \geqslant h(\mathbf{y}_{\rm n,:}, \xi_{\rm n,:}) = 
    e^{\mathbf{y}_{\rm n,:} \mathbf{t}_{\rm n,:}}\sigma(\xi_{\rm n,:})e^{-\frac{\mathbf{y}_{\rm n,:} + \xi_{\rm n,:}}{2} - \lambda(\xi_{\rm n,:})(\mathbf{y}_{\rm n,:}^2 - \xi_{\rm n,:}^2)},
\end{equation}
hence, we can rewrite Eq. (\ref{eq:E_y_ant}) as:
\begin{equation}
\label{eq:E_y}
    \ln q(\mathbf{Y}) = \E\left[\ln h(\mathbf{y}_{\rm n,:}, \xi_{\rm n,:}) + \ln p(\mathbf{Y}|\mathbf{Z}, \mathbf{V},\psi)\right].
\end{equation}

Hence, we can develop the first term as:
\begin{equation}
\label{eq:term1_y}
\begin{split}
    \ln h(\mathbf{y}_{\rm n,:}, \xi_{\rm n,:}) & = \sumn\sumc(\ln(\sigma(\xi_{\rm n,c}))) + y_{\rm n,c} t_{\rm n,c} - \unmed(y_{\rm n,c} + \xi_{\rm n,c})) - \lambda(\xi_{\rm n,c}))(y_{\rm n,c}^2 - \xi_{\rm n,c}^2))) \\
    & = \sumn\sumc(y_{\rm n,c} t_{\rm n,c} - \unmed y_{\rm n,c} - y_{\rm n,c}^2\lambda(\xi_{\rm n,c})) + \const \\
    & = \sumn((\mathbf{t}_{\rm n,:} - F_{\unmed})\Yn - \Yn \Lambda_{\xi_{\rm n,:}}\YnT) + \const.,
\end{split}
\end{equation}
where $F_{\unmed} \in \R^{C}$ is a row vector filled by $\unmed$ at all the positions. Also, we can develop the second term as follows:
\begin{equation}
\label{eq:term2_y}
\begin{split}
    \ln p(\mathbf{Y}|\mathbf{Z}, \mathbf{V},\psi) & = \sum_{n=1}^N \ln \N(\mathbf{z}_{n,:}\mathbf{V}^\top,\psi) \\
    & = \sum_{n=1}^{N}\left(-\unmed\lnpi + \unmed\ln \psi -\frac{\psi}{2}(\yn - \mathbf{z}_{n,:}\mathbf{V}^\top)(\Yn - \mathbf{z}_{n,:}\mathbf{V}^\top)^\top\right) \\
    & = \sum_{n=1}^{N}\left(-\frac{\psi}{2}\left(\Yn\YnT -2\Zn\VT\YnT\right)\right) + \const.
\end{split}
\end{equation}

Now, if we fill Eq. (\ref{eq:E_y}) with Eq. (\ref{eq:term1_y}) and (\ref{eq:term2_y}) and apply expectation, we obtain:
\begin{equation}
    q(\Yn) = \N(\langle \mathbf{y}_{\rm n,:}\rangle,\Sigma_{\mathbf{y}_{\rm n,:}}),
\end{equation}
where
\begin{align}
    \Sigma_{\mathbf{y}_{\rm n,:}}^{-1} & = \langle\psi\rangle\Ic + 2\Lambda_{\xi_{\rm n,:}} \\ \langle\mathbf{y}_{\rm n,:}\rangle & = \left(\mathbf{t}_{\rm n,:} - \unmed + \langle\psi\rangle\langle\mathbf{z}_{\rm n,:}\rangle\langle\mathbf{V}^\top\rangle\right)\Sigma_{\mathbf{y}_{\rm n,:}}.
\end{align}

Finally, to calculate the variational parameter $\bm{\xi}$, we have to maximize $L(q)$. In this case, we only have to take into account the term that includes $\bm{\xi}$ ($\ln(h(\mathbf{Y}, \bm{\xi}))$), defined as
\begin{equation}
\label{eq:elbo_dc}
    \E_q(\ln(h(\mathbf{Y}, \bm{\xi}))) = \sumn\sumc(\ln(\sigma(\xi_{\rm n,c})) + \langle y_{\rm n,c}\rangle t_{\rm n,c} - \unmed(\langle y_{\rm n,c}\rangle + \xi_{\rm n,c}) - \lambda(\xi_{\rm n,c})(\E[\y_{\rm n,c}^2] - \xi_{\rm n,c}^2)).
\end{equation}

Hence, to find the maximum value, we can derivate $\E_q(\ln(h(y_{\rm n,c}, \xi_{\rm n,c})))$ with respect to each $\xi_{\rm n,c}$ and equal to zero:
\begin{equation}
    \frac{\partial\E_q(\ln(h(y_{\rm n,c}, \xi_{\rm n,c})))}{\partial\xi_{\rm n,c}} = \lambda^{'}(\xi_{\rm n,c})(\E[y_{\rm n,c}^2]-\xi_{\rm n,c}^2) = 0,
\end{equation}
where the derivative function $\lambda^{'}(\xi_{\rm n,c})$ is monotonic for $\xi_{\rm n,c}   \geqslant 0$, and we can focus on nonnegative values of $\xi_{\rm n,c}$ due to its symmetry around 0:
\begin{equation}
    \lambda^{'}(\xi_{\rm n,c}) \neq 0 \longrightarrow \xi_{\rm n,c}^{new2} = \E[y_{\rm n,c}^2] = \langle y_{\rm n,c}^2 \rangle +\Sigma_{y_{\rm n,c}}.
\end{equation}
\newpage
\appendix
\setcounter{equation}{0}
\setcounter{figure}{0}
\setcounter{table}{0}
\setcounter{section}{2}
\makeatletter
\renewcommand{\theequation}{C.\arabic{equation}}
\renewcommand{\thefigure}{C\arabic{figure}}

\section{Lower Bound of BALDUR}
\label{sec:LB_BALDUR}

To calculate the final $L(q)$, we will focus on the formula defined in Eq. (\ref{eq:elboo}). That is:
\begin{equation}
\label{eq:elbo_BALDUR_APP}
\begin{split}
    L(q) & = -\int q(\bm{\Theta}) \ln\left( \frac{q(\bm{\Theta})}{p(\bm{\Theta},\mathbf{t},\mathbf{X}^{\M})}\right) d\bm{\Theta} = \int \prod_{i}q_i\left[\sum_i \ln(q_i) - \ln(p(\bm{\Theta},\mathbf{t},\X^{\M}))\right]d\bm{\Theta}\\
    & = \E_{q}\left[\ln (q(\bm{\Theta}))\right] - \E_{q}\left[\ln (p(\bm{\Theta}, \mathbf{t}, \mathbf{X}^{\M}))\right],
\end{split}
\end{equation}
where $\E_{q}\left[\ln (p(\bm{\Theta}, \mathbf{t}, \mathbf{X}^{\M}))\right]$ is the expectation of the joint posterior distribution w.r.t. $q(\mathbf{\Theta})$ and $\E_{q}[\ln 
 q(\bm{\Theta})]$ the entropy of $q(\mathbf{\Theta})$.

\subsection{Terms associated with $\E_{q}\left[\ln (p(\Theta, \mathbf{t}, \mathbf{X}^{\M}))\right]$}

This term comprises the following elements:
\begin{equation}
\begin{split}
\label{eq:full_elbo}
    &\E_{q}\left[\ln (p(\bm{\Theta}, \mathbf{t}, \mathbf{X}^{\M}))\right] = \E\Big[\ln p(\Y\vert\Z,\V,\psi) + \ln p(\mathbf{T}\vert\mathbf{Y}) + \ln p(\V\vert\omega) + \ln p(\psi) + \ln p(\omega) \\
    &+ \ln p(\mathbf{Z}|\mathbf{W}^{\M}\mathbbm{1}(s^\M = 0),\mathbf{A}^{\M}\mathbbm{1}(s^\M = 1),\tau) + \summ\Big(p(\mathbf{W}^{(m)}|\boldsymbol{\gamma}^{(m)}, \boldsymbol{\delta}^{(m)})^{\mathbbm{1}(s^{(m)} = 0)}
    \\ & p(\mathbf{\tilde{X}}^{(m)\top}\mathbf{A}^{(m)}|\boldsymbol{\gamma}^{(m)}, \boldsymbol{\delta}^{(m)})^{\mathbbm{1}(s^{(m)} = 1)} 
    + \ln p(\boldsymbol{\delta}^{(m)}) + \ln p(\boldsymbol{\gamma})^{(m)}\Big) + \ln p(\tau)\Big]
\end{split}
\end{equation}

Hence, we will separately analyze each term to subsequently merge them within Eq. (\ref{eq:full_elbo}). To do so, we will develop each term and subsequently apply the expectation over all the variables.

First, the term $\ln p(\mathbf{Z}|\mathbf{W}^{\M}\mathbbm{1}(s^\M = 0),\mathbf{A}^{\M}\mathbbm{1}(s^\M = 1),\tau)$ can be developed as:
\begin{equation}
\label{eq:full_elbo_z}
\begin{split}
    & \ln p(\mathbf{Z}|\mathbf{W}^{\M}\mathbbm{1}(s^\M = 0),\mathbf{A}^{\M}\mathbbm{1}(s^\M = 1),\tau) = -\frac{N}{2}\lnpi + \frac{N}{2}\ln \vert\tau\vert \\ 
    & - \sumn\unmed\tau\Bigg(\Zn -\summ\Bigg(\Big(\Xnm\WmT\Big)\mathbbm{1}(s^{(m)} = 0)+ \Big(\Xnm\Xsmt\Am\Big)\mathbbm{1}(s^{(m)} = 1)\Bigg)\Bigg)\\
    & \Bigg(\Zn -\summ\Bigg(\Big(\Xnm\WmT\Big)\mathbbm{1}(s^{(m)} = 0)+ \Big(\Xnm\Xsmt\Am\Big)\mathbbm{1}(s^{(m)} = 1)\Bigg)\Bigg)^\top\Bigg) .
\end{split}
\end{equation}

If we focus on the second part of the last equation, where the summation over $N$ starts, we can develop it as:
\begin{equation}
\begin{split}
    &\sumn\unmed\tau\Bigg(\Zn -\summ\Bigg(\Big(\Xnm\WmT\Big)\mathbbm{1}(s^{(m)} = 0)+ \Big(\Xnm\Xsmt\Am\Big)\mathbbm{1}(s^{(m)} = 1)\Bigg)\Bigg)\\
    & \Bigg(\Zn -\summ\Bigg(\Big(\Xnm\WmT\Big)\mathbbm{1}(s^{(m)} = 0)+ \Big(\Xnm\Xsmt\Am\Big)\mathbbm{1}(s^{(m)} = 1)\Bigg)\Bigg)^\top\Bigg) \\
    & =  \sumn\unmed\tau\Bigg(\Zn\ZnT - \Zn\mathbf{U}_{\rm n,:}^\top - \mathbf{U}_{\rm n,:}\mathbf{U}_{\rm n,:}^\top - \mathbf{U}_{\rm n,:}\ZnT\Bigg) \\
    & = \sumn\unmed\tau\Bigg(\Zn\ZnT - 2\Zn\mathbf{U}_{\rm n,:}^\top - \mathbf{U}_{\rm n,:}\mathbf{U}_{\rm n,:}^\top\Bigg),
\end{split}
\end{equation}
where 
\begin{equation}
\mathbf{U}_{\rm n,:} = \Bigg(\Zn -\summ\Bigg(\Big(\Xnm\WmT\Big)\mathbbm{1}(s^{(m)} = 0)+ \Big(\Xnm\Xsmt\Am\Big)\mathbbm{1}(s^{(m)} = 1)\Bigg)\Bigg).
\end{equation}

Now, if operate and return to Eq. (\ref{eq:full_elbo_z}), we can apply the expectation over the model variables and obtain:
\begin{equation}
\label{eq:casi_fin_z}
\begin{split}
    & \E\left[\ln p(\mathbf{Z}|\mathbf{W}^{\M}\mathbbm{1}(s^\M = 0),\mathbf{A}^{\M}\mathbbm{1}(s^\M = 1),\tau)\right] = -\frac{N}{2}\lnpi + \frac{N}{2}\E\left[\ln \vert\tau\vert\right]  \\
    & -2\langle\Zn\rangle\E\left[\mathbf{U}_{\rm n,:}\right] + \summ\left[\Xnm\langle\WmT\Wm\rangle\XnmT\right]\mathbbm{1}(s^{(m)} = 0) \\
    & + \summ\left[\Xnm\Xsmt\langle\Am\AmT\rangle\Xsm\XnmT\right]\mathbbm{1}(s^{(m)} = 1) \\
    & + \summ\sum_{m'\neq m}^\M \Bigg(\left[\Xnm\langle\WmT\rangle\langle\mathbf{W}^{(m')}\rangle\mathbf{x}_{\rm n,:}^{(m')\top}\right]\mathbbm{1}(s^{(m)} = 0) \\
    & + \left[\Xnm\Xsmt\langle\Am\rangle\langle\mathbf{A}^{(m')\top}\rangle\mathbf{\tilde{X}}^{(m')}\mathbf{x}_{\rm n,:}^{(m')\top}\right]\mathbbm{1}(s^{(m)} = 1)\Bigg).
\end{split}
\end{equation}

Also note that $\E\left[\ln(\tau)\right] = dg(\alpha^\tau) - \ln \beta^\tau$, and $dg()$ is the digamma function and $\E\left[\tau\right] = \frac{\alpha^\tau}{\beta^\tau}$.

Hence, we can simplify Eq. (\ref{eq:casi_fin_z}) as:
\begin{equation}
\begin{split}
    \E\left[\ln p(\mathbf{Z}|\mathbf{W}^{\M}\mathbbm{1}(s^\M = 0),\mathbf{A}^{\M}\mathbbm{1}(s^\M = 1),\tau)\right] = & -\frac{N}{2}\lnpi + \frac{N}{2}\left(dg(\alpha^\tau) - \ln \beta^\tau\right) \\ &-\unmed\frac{\alpha^\tau}{\beta^\tau}\left(\beta^\tau - \beta^\tau_0\right).
\end{split}
\end{equation}

Now, the development of $\ln p(\Y\vert\Z,\V,\psi)$ will be analog to $\ln p(\mathbf{Z}|\mathbf{W}^{\M}\mathbbm{1}(s^\M = 0),\mathbf{A}^{\M}\mathbbm{1}(s^\M = 1),\tau)$. That is, we can develop the output term as:
\begin{equation}
    \E\left[\ln p(\Y\vert\Z,\V,\psi)\right] = -\frac{N}{2}\lnpi + \frac{N}{2}\left(dg(\alpha^\psi) - \ln \beta^\psi\right) -\unmed\frac{\alpha^\psi}{\beta^\psi}\left(\beta^\psi - \beta^\psi_0\right).
\end{equation}

Also, if we focus in $\ln p(\mathbf{T}\vert\mathbf{Y})$, we will have to apply the expectation over the logarithm of the lower bounded expression depicted in Eq. (\ref{eq:elbo_dc}). Thus, we will finally achieve:
\begin{equation}
    \E(\ln(h(\mathbf{Y}, \bm{\xi}))) = \sumn\sumc(\ln(\sigma(\xi_{\rm n,c})) + \langle y_{\rm n,c}\rangle t_{\rm n,c} - \unmed(\langle y_{\rm n,c}\rangle + \xi_{\rm n,c}) - \lambda(\xi_{\rm n,c})(\langle\y_{\rm n,c}^2\rangle - \xi_{\rm n,c}^2)).
\end{equation}

Furthermore, we can develop the expectation over the output noise $\psi$ as:
\begin{equation}
\label{eq:psi_jon}
\begin{split}
    & \E\left[\ln p(\psi)\right] = \E\left[\alpha_0^\psi\ln\beta_0^\psi - \ln\left(\Gamma(\alpha_0^\psi)\right) - \beta_0^\psi \psi + (\alpha_0^\psi -1)\ln(\tau)\right] \\
    & = \alpha_0^\psi\ln\beta_0^\psi - \ln\left(\Gamma(\alpha_0^\psi)\right) - \beta_0^\psi \frac{\alpha_0^\psi}{\beta_0^\psi} + (\alpha_0^\psi -1)(dg(\alpha_0^\psi) - \ln \beta_0^\psi).
\end{split}
\end{equation}

Moreover, we can develop the $\boldsymbol{\omega}$ following a similar procedure as in Eq. (\ref{eq:psi_jon}). That is, 
\begin{equation}
\label{eq:omegaa}
    \E\left[\boldsymbol{\omega}\right] = K\left(\alpha_0^\omega\ln \beta_0^\omega - \ln \left(\Gamma(\alpha_0^\omega)\right)\right) + \sumk\left(-\beta_0^\omega\frac{\alpha_k^\omega}{\beta_k^\omega} + (\alpha_0^\omega-1)(dg(\alpha_k^\omega) - \ln \beta_k^\omega)\right)
\end{equation}

Finally, regarding the output view, we can develop $\ln p(\mathbf{V}\vert\boldsymbol{\omega})$ as follows:
\begin{equation}
\begin{split}
    &\E\left[\ln p(\mathbf{V}\vert\boldsymbol{\omega})\right] = \E\left[\sumk\left(-\unmed\lnpi + \unmed\ln\omega_k - \frac{\omega_k}{2}\VkT\Vk\right)\right] \\
    & = -\frac{K}{2}\lnpi + \unmed\sumk\E\left[\ln \omega_k\right] - \unmed\sumk\frac{\alpha_k^\omega}{\beta_k^\omega}diag(\langle\V\VT\rangle) \\
    & = -\frac{K}{2}\lnpi + \unmed\sumk dg(\alpha_k^\omega) - \sumk\alpha_k^\omega + \beta_0^\omega \sumk\frac{\alpha_k^\omega}{\beta_k^\omega}
\end{split}
\end{equation}

To calculate the expectation over the input noise $\tau$, we will follow the same pattern as in Eq. (\ref{eq:psi_jon}). That is:
\begin{equation}
\begin{split}
    & \E\left[\ln p(\tau)\right] = \E\left[\alpha_0^\tau\ln\beta_0^\tau - \ln\left(\Gamma(\alpha_0^\tau)\right) - \beta_0^\tau \tau + (\alpha_0^\tau -1)\ln(\tau)\right] \\
    & = \alpha_0^\tau\ln\beta_0^\tau - \ln\left(\Gamma(\alpha_0^\tau)\right) - \beta_0^\tau \frac{\alpha_0^\tau}{\beta_0^\tau} + (\alpha_0^\tau -1)(dg(\alpha_0^\tau) - \ln \beta_0^\tau).
\end{split}
\end{equation}

Also, in an analog way as in Eq. (\ref{eq:omegaa}), we can develop the expectations over both $\boldsymbol{\delta}$ and $\boldsymbol{\gamma}$ as:
\begin{equation}
\label{eq:alphaaa}
    \E\left[\boldsymbol{\delta}\right] = K\left(\alpha_0^\delta\ln \beta_0^\delta - \ln \left(\Gamma(\alpha_0^\delta)\right)\right) + \sumk\left(-\beta_0^\delta\frac{\alpha_k^\delta}{\beta_k^\delta} + (\alpha_0^\delta-1)(dg(\alpha_k^\delta) - \ln \beta_k^\delta)\right),
\end{equation}
and
\begin{equation}
\label{eq:gamaaa}
    \E\left[\boldsymbol{\gamma}\right] = D\left(\alpha_0^\gamma\ln \beta_0^\gamma - \ln \left(\Gamma(\alpha_0^\gamma)\right)\right) + \sumd\left(-\beta_0^\gamma\frac{\alpha_d^\gamma}{\beta_d^\gamma} + (\alpha_0^\gamma-1)(dg(\alpha_d^\gamma) - \ln \beta_d^\gamma)\right),
\end{equation}
respectively. Note that both Eq. (\ref{eq:alphaaa}) and (\ref{eq:gamaaa}) are a particular case for a specific $m$ view.

Finally, we can directly go over the weight variables, i.e., $\mathbf{W}$ or $\mathbf{A}$ depending on whether we will work over the primal or dual space, respectively. Thus, as the developments are analog, we will first develop the primal case for the $m$-th view as:
\begin{equation}
\begin{split}
\label{eq:www}
    &\ln p(\mathbf{W}^{(m)}\vert \boldsymbol{\delta}^{(m)}, \boldsymbol{\gamma}^{(m)}) = -\frac{K D}{2}\lnpi + \unmed\sumk\sumd(\ln(\gamma_d\delta_k)) - \unmed\sumk\sumd(w_{\rm k,d}^2\delta_k\gamma_d) \\
    & = -\frac{K D}{2}\lnpi + \frac{D}{2}\sumk\ln\delta_k + \frac{K}{2}\sumd\ln\gamma_d - \unmed\sumk\sumd(w_{\rm k,d}^2\delta_k\gamma_d). 
\end{split}
\end{equation}

Now, if we apply the expectation over Eq. (\ref{eq:www}), we obtain:
\begin{equation}
\begin{split}
    &\E\left[\ln p(\mathbf{W}^{(m)}\vert \boldsymbol{\delta}^{(m)}, \boldsymbol{\gamma}^{(m)})\right] = -\frac{K D}{2}\lnpi + \frac{D}{2}\sumk(dg(\alpha_k^\delta) - \ln \beta_k^\delta) + \frac{K}{2}\sumd(dg(\alpha_d^\gamma) - \ln \beta_d^\gamma) \\
    & - \unmed\sumk\sumd\left[\frac{\alpha_k^\delta\alpha_d^\gamma}{\beta_k^\delta\beta_d^\gamma}\langle w_{\rm k,d}^2\rangle\right].
\end{split}
\end{equation}

Finally, the dual case will be analog:
\begin{equation}
\begin{split}
    &\E\left[\ln p(\mathbf{A}^{(m)}\mathbf{\tilde{X}}^{(m)}\vert \boldsymbol{\delta}^{(m)}, \boldsymbol{\gamma}^{(m)})\right] = -\frac{K D}{2}\lnpi + \frac{D}{2}\sumk(dg(\alpha_k^\delta) - \ln \beta_k^\delta) \\ &+ \frac{K}{2}\sumd(dg(\alpha_d^\gamma) - \ln \beta_d^\gamma) 
    - \unmed\sumk\sumd\left[\frac{\alpha_k^\delta\alpha_d^\gamma}{\beta_k^\delta\beta_d^\gamma}Tr(\mathbf{\tilde{X}}^{(m)}\mathbf{\tilde{X}}^{(m)\top}\langle\mathbf{a}_{\rm :,k}\mathbf{a}_{\rm :,k}^\top\rangle)\right].
\end{split}
\end{equation}

\subsection{Terms associated to $\E_{q}[\ln q(\Theta)]$}

the entropy term will look like:
\begin{equation}
\begin{split}
    &\E_{q}[\ln q(\boldsymbol{\Theta})] = \E\Bigg[\ln q(\mathbf{Y}) + \ln q(\mathbf{V}) + \ln q(\boldsymbol{\omega}) + \ln q(\psi) + \ln q(\mathbf{Z})\\ 
    & + \summ\Big[\ln q(\mathbf{W}^{(m)})\mathbbm{1}(s^{(m)} = 0) + \ln q(\mathbf{A}^{(m)})\mathbbm{1}(s^{(m)} = 1) \\ &+ \ln q(\boldsymbol{\delta}^{(m)}) + \ln q(\boldsymbol{\gamma}^{(m)})\Big] + \ln q(\tau)\Bigg].
\end{split}
\end{equation}

Also, as in $\E_{q}\left[\ln (p(\boldsymbol{\Theta}, \mathbf{t}, \mathbf{X}^{\M}))\right]$, each element can be analyzed in an independent way as:
\begin{align}
    \ln q(\mathbf{Y}) & = \sumn\left(\unmed\ln 2\pi e + \unmed\ln \vert\Sigma_{\mathbf{y}_{\rm n,:}}\vert\right) = \frac{N}{2}\ln 2\pi e + \frac{N}{2}\ln \vert\Sigma_{\mathbf{Y}}\vert\\
    \ln q(\mathbf{V}) & = \sumc\left(\unmed\ln 2\pi e + \unmed\ln \vert\Sigma_{\mathbf{V}}\vert\right) = \frac{C}{2}\ln 2\pi e + \frac{C}{2}\ln \vert\Sigma_{\mathbf{V}}\vert \\
    \ln q(\boldsymbol{\omega}) & = \sumk\left(\alpha_k^\omega + \ln \Gamma(\alpha_k^\omega) - (1-\alpha_k^\omega)dg(\alpha_k^\omega) - \ln \beta_k^\omega\right) \\
    \ln q(\boldsymbol{\delta^{(m)}}) & = \sumk\left(\alpha_k^{\delta^{(m)}} + \ln \Gamma(\alpha_k^{\delta^{(m)}}) - (1-\alpha_k^{\delta^{(m)}})dg(\alpha_k^{\delta^{(m)}}) - \ln \beta_k^{\delta^{(m)}}\right) \\
    \ln q(\boldsymbol{\gamma^{(m)}}) & = \sumd\left(\alpha_d^{\gamma^{(m)}} + \ln \Gamma(\alpha_d^{\gamma^{(m)}}) - (1-\alpha_d^{\gamma^{(m)}})dg(\alpha_d^{\gamma^{(m)}}) - \ln \beta_d^{\gamma^{(m)}}\right) \\
    \ln q(\psi) & = \alpha^\psi + \ln \Gamma(\alpha^\psi) - (1-\alpha^\psi)dg(\alpha^\psi) - \ln \beta^\psi \\
    \ln q(\tau) & = \alpha^\tau + \ln \Gamma(\alpha^\tau) - (1-\alpha^\tau)dg(\alpha^\tau) - \ln \beta^\tau \\
    \ln q(\mathbf{Z}) &= \frac{N}{2}\ln 2\pi e + \frac{N}{2}\ln\vert\Sigma_{\mathbf{Z}}\vert\\
    \ln q(\mathbf{W}^{(m)}) &= \frac{K}{2}\ln 2\pi e + \unmed\sumk\ln\vert\Sigma_{\mathbf{W}^{(m)}_{\rm k,:}}\vert\\
    \ln q(\mathbf{A}^{(m)}) &= \frac{K}{2}\ln 2\pi e + \unmed\sumk\ln\vert\Sigma_{\mathbf{a}^{(m)}_{\rm :,k}}\vert.
\end{align}

\subsection{Complete Lower Bound}

Finally, we can join both elements of $\E_{q}\left[\ln (p(\bm{\Theta}, \mathbf{t}, \mathbf{X}^{\M}))\right]$ and $\E_{q}[\ln q(\bm{\Theta})]$ within Eq. (\ref{eq:elbo_BALDUR_APP}) to ontain the final $L(q)$ as:
\begin{equation}
\label{eq:Lq}
    \begin{split}
        L(q) = &\frac{\N}{2}\ln |\Sigma_{\mathbf{Z}}|- (2 + \frac{\N}{2} - \alpha_0^\tau)\ln(\beta^\tau) + \frac{\N}{2}\ln|\Sigma_{\mathbf{y}}| + \frac{C}{2}\ln|\Sigma_{\mathbf{V}}| + (\alpha_0^\omega -2)\sum_{\rm k}^K\ln(\beta_{\rm k}^\omega) \\ & - (2 + \frac{\N}{2} - \alpha_0^\psi)\ln(\beta^{\psi}) + \sum_m^M\bigg[\left(\frac{K}{2} + \alpha_0^{\gamma^{\rm (m)}} -2\right)\sum_{\rm d}^{D^{\rm (m)}}\ln(\beta_{\rm d}^{\gamma^{\rm (m)}}) \\ & + \left(\frac{K}{2} + \alpha_0^{\delta^{\rm (m)}} -2\right)\sum_{\rm k}^K\ln(\beta_{\rm k}^{\delta^{\rm (m)}}) + \sum_{\rm d}^{D^{\rm (m)}}\left(\beta_0^{\gamma^{\rm (m)}}\frac{\alpha_{\rm d}^{\gamma^{\rm (m)}}}{\beta_{\rm d}^{\gamma^{\rm (m)}}}\right) + \sum_{\rm k}^K\left(\beta_0^{\delta^{\rm (m)}}\frac{\alpha_{\rm k}^{\delta^{\rm (m)}}}{\beta_{\rm k}^{\delta^{\rm (m)}}}\right) \\ & + \unmed\sum_{\rm k}^K\sum_{\rm d}^{D^{\rm (m)}}\left(\frac{\alpha_{\rm k}^{\delta^{\rm (m)}} \alpha_{\rm d}^{\gamma^{\rm (m)}}}{\beta_{\rm k}^{\delta^{\rm (m)}} \beta_{\rm d}^{\gamma^{\rm (m)}}}\langle\Wkd\rangle\right)\mathbbm{1}(s^{\rm (m)} = 0) \\ & + \unmed\sum_{\rm k}^K\sum_{\rm d}^{D^{\rm (m)}}\left(\frac{\alpha_{\rm k}^{\delta^{\rm (m)}} \alpha_{\rm d}^{\gamma^{\rm (m)}}}{\beta_{\rm k}^{\delta^{\rm (m)}} \beta_{\rm d}^{\gamma^{\rm (m)}}}Tr\left(\Xsm\Xsmt\langle\mathbf{a}_{:,\rm k}\mathbf{a}_{:,\rm k}^T\rangle\right)\mathbbm{1}(s^{\rm (m)} = 1)\right) \bigg] \\ & - \sum_n^N\left(\ln(\sigma(\xi_{\rm n,:})) + \langle\mathbf{y}_{\rm n,:}\rangle\mathbf{t}_{\rm n,:}^\top - \unmed(\langle\mathbf{y}_{\rm n,:}\rangle + \xi_{\rm n,:}) - \lambda(\xi_{\rm n,:})(\langle\mathbf{y}_{\rm n,:}^2\rangle - \xi_{\rm n,:}^2)\right).
    \end{split}
\end{equation}

\newpage
\bibliographystyle{plain}
\bibliography{refs_BALDUR.bib}

\end{document}